\documentclass{article}
\usepackage[preprint]{icml2026}
\usepackage{microtype}
\usepackage{graphicx}
\usepackage{subcaption}
\usepackage{booktabs} 

\usepackage{hyperref}

\usepackage{icml2026}

\usepackage{amsmath}
\usepackage{amssymb}
\usepackage{mathtools}
\usepackage{amsthm}

\usepackage[capitalize,noabbrev]{cleveref}

\usepackage{graphicx}
\usepackage{subcaption}
\usepackage{array}
\usepackage{bbm}
\usepackage{adjustbox}
\usepackage{paralist}
\usepackage{pifont}       
\usepackage{bbding}       
\usepackage{fontawesome}  
\usepackage{colortbl}
\definecolor{blue}{RGB}{218,232,252}
\definecolor{green}{RGB}{112, 173, 71}
\definecolor{redbrown}{RGB}{255 127 80}
\definecolor{aliceblue}{rgb}{0.94, 0.97, 1.0}
\usepackage{multirow}

\usepackage{hyperref}

\newcommand{\redtoken}[1]{\texttt{\textcolor{redbrown}{#1}}}
\newcommand{\greentoken}[1]{\texttt{\textcolor{green}{#1}}}

\usepackage{subcaption}  

\theoremstyle{plain}

\theoremstyle{definition}

\theoremstyle{remark}

\usepackage[textsize=tiny]{todonotes}

\icmltitlerunning{Submission and Formatting Instructions for ICML 2026}

\begin{document}

\twocolumn[
  \icmltitle{LLMBind: A Unified Modality-Task Integration Framework}



  \icmlsetsymbol{equal}{*}



    \begin{icmlauthorlist}

\icmlauthor{Bin Zhu}{pku}
\icmlauthor{Munan Ning}{pku,pcl}
\icmlauthor{Peng Jin}{pku}
\icmlauthor{Bin Lin}{pku}
\icmlauthor{Jinfa Huang}{ur}
\icmlauthor{Qi Song}{hkbu}
\icmlauthor{Junwu Zhang}{pku}
\icmlauthor{Zhenyu Tang}{pku}
\icmlauthor{Mingjun Pan}{pku}
\icmlauthor{Li Yuan}{pku}

\end{icmlauthorlist}

\icmlaffiliation{pku}{Peking University, China}
\icmlaffiliation{pcl}{Peng Cheng Laboratory, China}
\icmlaffiliation{ur}{University of Rochester, USA}
\icmlaffiliation{hkbu}{Hong Kong Baptist University, China}

\icmlcorrespondingauthor{Li Yuan}{PKU}

  \icmlkeywords{Machine Learning, ICML}

  \vskip 0.3in
]



\printAffiliationsAndNotice{}  

\begin{abstract}
Despite recent progress in Multi-Modal Large Language Models (MLLMs), it remains challenging to integrate diverse tasks ranging from pixel-level perception to high-fidelity generation. Existing approaches often suffer from either restricted task extensibility or severe performance degradation due to modality interference.
n this paper, we present \textbf{LLMBind}, an extensible framework that unifies multimodal tasks through a dual-pathway mechanism: \greentoken{ \textit{In-Situ}} semantic embeddings for localization-sensitive tasks like semantic segmentation and \redtoken{\textit{Ex-Situ}} task-prompts for generation across image, video, and audio modalities. Additionally, we employ a Mixture-of-Experts (MoE) architecture to route task-specific tokens, thereby achieving modality disentanglement and mitigating negative transfer. We also curate a 400k multi-turn interactive dataset focused on iterative visual refinement to enable human-like interaction. 
Extensive experiments demonstrate that LLMBind achieves excellent performance across multiple perception and generation benchmarks while maintaining superior expandability. 


\end{abstract}

\section{Introduction}
\label{sec:intro}

\begin{figure}[ht]
\centering
    \includegraphics[width=1.0\linewidth]{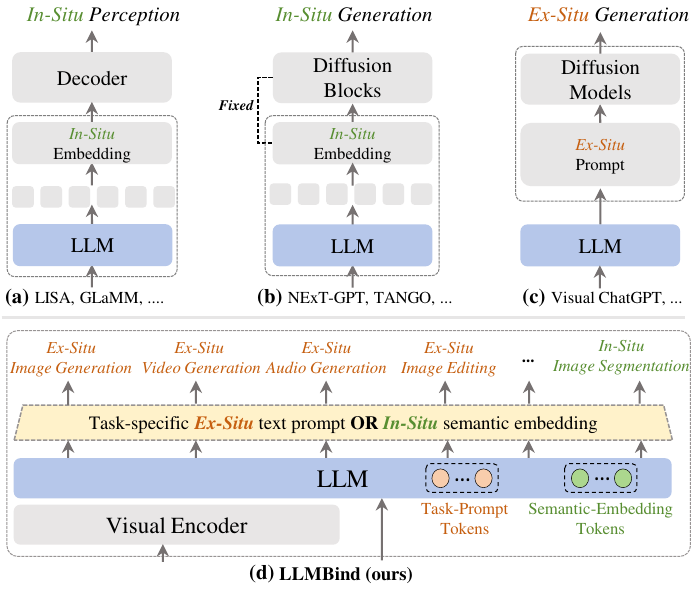}  
\caption{Existing frameworks of multimodal perception and generation. Unlike previous works that rely on either \redtoken{ \textit{Ex-Situ}} textual prompts or \greentoken{ \textit{In-Situ}} rigid representation coupling, LLMBind introduces a task-specific hybrid strategy. By decoupling generation from perception through distinct token pathways, LLMBind preserves fine-grained grounding while enabling plug-and-play generative flexibility.}
\label{fig:intro}
\end{figure}

Artificial Intelligence is transitioning from specialized models toward general-purpose multimodal agents. 
While Vision-Language Models (VLMs)~\cite{lai2023lisa, rasheed2024glamm} excel in fine-grained perception and diffusion models~\cite{NicholDRSMMSC22, RombachBLEO22} provide high-fidelity synthesis. integrating both capabilities into a single framework is difficult.
This difficulty stems from a fundamental architectural mismatch: perception requires tightly grounded \textit{In-Situ} representations for spatial precision, whereas scalable generation favors \textit{Ex-Situ} modular designs for flexibility and extensibility. Consequently, current approaches often specialize in one domain rather than achieving effective integration.

As depicted in \cref{fig:intro} (a), existing perception models like RexSeek~\cite{jiang2025referring}, LISA~\cite{lai2023lisa}, PixelLM~\cite{ren2024pixellm}, PSALM~\cite{zhang2024psalm} and GLaMM~\cite{rasheed2024glamm}, embed visual signals directly into the latent space of the LLM. These \textit{In-Situ} latent embeddings support high-precision tasks like referring segmentation, but the models lack the ability for content generation. 
Generation paradigms further divide into two categories. \textit{In-Situ} generation approaches, as depicted in \cref{fig:intro} (b), such as NExT-GPT~\cite{wu2023next},  RISE-T2V~\cite{zhang2025rise}, Sta-v2a~\cite{ren2025sta}, GILL~\cite{koh2023generating}, DreamLLM~\cite{dong2023dreamllm}, MiniGPT5~\cite{zheng2023minigpt} and SEED-X~\cite{ge2024seed}, align LLM embeddings with generative latent spaces to enable synthesis. However, However, this creates a rigid coupling that restricts flexibility and complicates the adoption of new generative backbones.
Conversely, the pure \textit{Ex-Situ} generation, as shown in \cref{fig:intro} (c),  including systems like  Omnigen~\cite{xiao2025omnigen}  Visual ChatGPT~\cite{wu2023visual}, Creatilayout~\cite{zhang2025creatilayout} AudioGPT~\cite{huang2024audiogpt}, InternGPT~\cite{liu2023interngpt}, InstructPix2Pix~\cite{brooks2023instructpix2pix} and HuggingGPT~\cite{shen2023hugginggpt}, which invoke external tools via textual prompts. This offers strong modularity but suffers from a textual bottleneck, 
where sparse tokens cannot convey dense spatial information for pixel-level perception.
Furthermore, jointly optimizing dense perceptual and abstract generative objectives triggers task interference, which degrades overall performance. These limitations demonstrate that reconciling distinct architectural and objective choices within a single framework remains a fundamental difficulty.

To resolve these contradictions, we introduce LLMBind, as shown in \cref{fig:intro} (d), a hybrid framework designed to balance \textit{In-Situ} perception with \textit{Ex-Situ} generation. Our core philosophy is a strategic bifurcation of multimodal tasks based on their intrinsic requirements for perception and generation granularity. For perception tasks, we employ learnable \textbf{\greentoken{Semantic-Embedding Tokens}} to extract fine-grained visual information directly from the LLM’s latent space. Conversely, for generation-heavy tasks, we utilize \textbf{\redtoken{Task-Prompt Tokens}} to steer frozen, state-of-the-art foundation models. Such a decoupled architecture empowers LLMBind with seamless \textit{plug-and-play} expandability for emerging generative models without sacrificing the precision required for dense perception. 
To resolve optimization conflicts in shared-backbone multi-task learning, we propose a Task-Specific MoE architecture. The system routes task-specific tokens to dedicated LoRA experts to ensure explicit parameter-level isolation for different modalities. Consequently, LLMBind achieves functional decoupling between tasks, which effectively preserves task-specific expertise while leveraging the collective reasoning power of the LLM.

Furthermore, static instruction following does not meet the requirements of real-world collaboration. To solve this, we create an instruction-generative interleaved dataset with 400k samples. The dataset uses multi-turn formats to simulate refinement tasks, such as progressive image editing and multi-step visual generation. This allows LLMBind to achieve continuous visual refinement for human-like interaction. The main contributions are as follows:
\begin{itemize}
    \item We propose LLMBind, a dual-pathway framework unifying perception and generation via \textit{In-Situ} embedding and \textit{Ex-Situ} prompting.
    \item We introduce a Task-Specific MoE strategy to mitigate modality interference through parameter-level isolation during joint multi-task training.
    \item We release a 400k instruction-generative interleaved dataset for multi-turn visual refinement, including image editing and multi-step generation.
    \item LLMBind outperforms specialized models in segmentation and generation, with human evaluation confirming high alignment with user instructions.
\end{itemize}

\section{Related Work}

\subsection{Cross-Modal Generation and Editing}

Multi-modal generation has remained a highly active research area in recent years.
These models~\cite{ramesh2022hierarchical,rombach2022high,hong2022cogvideo,kong2020diffwave,yang2023diffsound,xiao2025omnigen,zhang2025creatilayout} allow the generation of corresponding images, videos, and audio based on specified text prompts. However, this approach requires users to directly write text prompts for image generation, lacking interactive capabilities. Another approach to cross-modal generation~\cite{wu2023next, zeqiang2023mini}, combines dialogue with generation, making it more user-friendly and aligned with current user needs. Given that generated images may not always meet user requirements, image editing has emerged as another active research area within the field of artificial intelligence-generated content. Representative works in this area include Text2live~\cite{bar2022text2live}, Imagic~\cite{kawar2023imagic}, Sdedit~\cite{meng2021sdedit}. 

\subsection{Cross-Modal  Segmentation}
With the introduction of frameworks such as LanguageBind~\cite{zhu2023languagebind} and ImageBind~\cite{girdhar2023imagebind}, which combine multi-modal data in a shared semantic space, several works focusing on cross-modal understanding~\cite{abs-2304-08485,abs-2304-10592, yin2023survey, jin2023chat} have emerged.
Along with these advancements, there has been a growing interest in text-guided image referring segmentation, a new approach that enables users to interact with images using user instruction. Unlike traditional semantic segmentation, which assigns class labels to pixels, referring segmentation~\cite{lai2023lisa, liu2023gres, zhang2024llava} aims to segment target objects based on complex textual instructions. 

\begin{table}
  \tabcolsep=0.1cm
  \caption{\textbf{Comparison of functionalities among different MLLMs.}\textcolor{green}{\ding{52}} indicates capability, \textcolor{red}{\ding{55}} indicates limitation, and ``\textbf{-}'' hints at the potential for easy extension. We cover a range of tasks including text conversation (Text-Con), image comprehension (Img-Com), image generation (Img-Gen), image editing (Img-Edit), video generation (Vid-Gen), audio generation (Aud-Gen), image segmentation (Img-Seg), and image detection (Img-Det).}
  \label{tab:mllm}
  \centering
  
  \begin{adjustbox}{scale=0.5}
  \begin{tabular}{lcccccccc}
    \toprule
    \textbf{MLLMs} & \textbf{\small{Text-Con}} & \textbf{\small{Img-Com}} & \textbf{\small{Img-Gen}} & \textbf{\small{Img-Edit}} & \textbf{\small{Vid-Gen}} & \textbf{\small{Aud-Gen}} & \textbf{\small{Img-Seg}} & \textbf{\small{Img-Det}}  \\
    \midrule
    LLaMA~\cite{abs-2302-13971} & \textcolor{green}{\ding{52}} & \textcolor{red}{\ding{55}} & \textcolor{red}{\ding{55}} & \textcolor{red}{\ding{55}} &\textcolor{red}{\ding{55}}  &\textcolor{red}{\ding{55}}  &\textcolor{red}{\ding{55}}  &\textcolor{red}{\ding{55}} \\
    Vicuna~\cite{vicuna} & \textcolor{green}{\ding{52}} & \textcolor{red}{\ding{55}} & \textcolor{red}{\ding{55}} & \textcolor{red}{\ding{55}} &\textcolor{red}{\ding{55}}  &\textcolor{red}{\ding{55}}  &\textcolor{red}{\ding{55}}  &\textcolor{red}{\ding{55}} \\
    Flaminggo~\cite{AlayracDLMBHLMM22} & \textcolor{green}{\ding{52}} & \textcolor{green}{\ding{52}} & \textcolor{red}{\ding{55}} & \textcolor{red}{\ding{55}} &\textcolor{red}{\ding{55}}  &\textcolor{red}{\ding{55}}  &\textcolor{red}{\ding{55}}  &\textcolor{red}{\ding{55}} \\
    BLIP2~\cite{li2023blip} & \textcolor{green}{\ding{52}} & \textcolor{green}{\ding{52}} & \textcolor{red}{\ding{55}} & \textcolor{red}{\ding{55}} &\textcolor{red}{\ding{55}}  &\textcolor{red}{\ding{55}}  &\textcolor{red}{\ding{55}}  &\textcolor{red}{\ding{55}} \\
    mPLUG-Owl~\cite{ye2023mplug} & \textcolor{green}{\ding{52}} & \textcolor{green}{\ding{52}} & \textcolor{red}{\ding{55}} & \textcolor{red}{\ding{55}} &\textcolor{red}{\ding{55}}  &\textcolor{red}{\ding{55}}  &\textcolor{red}{\ding{55}}  &\textcolor{red}{\ding{55}} \\
    DetGPT~\cite{pi2023detgpt} & \textcolor{green}{\ding{52}} & \textcolor{green}{\ding{52}} & \textcolor{red}{\ding{55}} & \textcolor{red}{\ding{55}} &\textcolor{red}{\ding{55}}  &\textcolor{red}{\ding{55}}  &\textcolor{red}{\ding{55}}  &\textcolor{green}{\ding{52}} \\
    LLaVA~\cite{abs-2304-08485} &   \textcolor{green}{\ding{52}} & \textcolor{green}{\ding{52}} & \textcolor{red}{\ding{55}} & \textcolor{red}{\ding{55}} &\textcolor{red}{\ding{55}}  &\textcolor{red}{\ding{55}}  &\textcolor{red}{\ding{55}}  &\textcolor{red}{\ding{55}} \\
    Mini-GPT4~\cite{abs-2304-10592} &   \textcolor{green}{\ding{52}} & \textcolor{green}{\ding{52}} & \textcolor{red}{\ding{55}} & \textcolor{red}{\ding{55}} &\textcolor{red}{\ding{55}}  &\textcolor{red}{\ding{55}}  &\textcolor{red}{\ding{55}}  &\textcolor{red}{\ding{55}} \\
    Audiogpt~\cite{huang2023audiogpt} &  \textcolor{green}{\ding{52}} & \textcolor{red}{\ding{55}} & \textcolor{red}{\ding{55}} &\textcolor{red}{\ding{55}}  &\textcolor{red}{\ding{55}} & \textcolor{green}{\ding{52}}  &\textcolor{red}{\ding{55}}  &\textcolor{red}{\ding{55}} \\
    
    Otter~\cite{li2023otter} & \textcolor{green}{\ding{52}} & \textcolor{green}{\ding{52}} & \textcolor{red}{\ding{55}} & \textcolor{red}{\ding{55}} &\textcolor{red}{\ding{55}}  &\textcolor{red}{\ding{55}}  &\textcolor{red}{\ding{55}}  &\textcolor{red}{\ding{55}} \\
    LISA~\cite{lai2023lisa}  & \textcolor{green}{\ding{52}} & \textcolor{green}{\ding{52}} & \textcolor{red}{\ding{55}} & \textcolor{red}{\ding{55}} &\textcolor{red}{\ding{55}}  &\textcolor{red}{\ding{55}}  &\textcolor{green}{\ding{52}}  &\textcolor{red}{\ding{55}} \\

    GLaMM~\cite{rasheed2023glamm}  & \textcolor{green}{\ding{52}} & \textcolor{green}{\ding{52}} & \textcolor{red}{\ding{55}} & \textcolor{red}{\ding{55}} &\textcolor{red}{\ding{55}}  &\textcolor{red}{\ding{55}}  &\textcolor{green}{\ding{52}}  &\textcolor{red}{\ding{55}} \\
    

    NExTGPT~\cite{wu2023next} & \textcolor{green}{\ding{52}} & \textcolor{green}{\ding{52}} & \textcolor{green}{\ding{52}} & \textcolor{red}{\ding{55}} & \textcolor{green}{\ding{52}}  &\textcolor{green}{\ding{52}}  &\textcolor{red}{\ding{55}}  &\textcolor{red}{\ding{55}} \\
    \midrule
    
    \rowcolor{aliceblue}\textbf{LLMBind (Ours)} & \textcolor{green}{\ding{52}} & \textcolor{green}{\ding{52}} & \textcolor{green}{\ding{52}} & \textcolor{green}{\ding{52}} & \textcolor{green}{\ding{52}} & \textcolor{green}{\ding{52}} & \textcolor{green}{\ding{52}} & \textbf{-} \\
    \bottomrule
  \end{tabular}
  \end{adjustbox}
\end{table}
\subsection{Mixture-of-Experts}
\label{author info}

The concept of Mixture-of-Experts (MoE) in artificial neural networks has evolved. Early methods activated all experts, leading to high computational requirements~\cite{eigen2013learning}. Subsequent research introduced sparse MoE models, where distinct experts are selectively activated using a learnable gating mechanism~\cite{shazeer2017outrageously, lepikhin2020scaling, fedus2022switch}. These approaches have shown success 
in computer vision~\cite{eigen2013learning, ahmed2016network, gross2017hard, wang2020deep} and natural language processing~\cite{shazeer2017outrageously, lepikhin2020scaling, zhou2022mixture} domains.

\subsection{Multimodal Large Language Models}

Recently, the emergence of Large Language Models (LLM) has ushered AI into a new era. With the open-sourcing of LLM models like LLaMA~\cite{abs-2302-13971} and Vicuna~\cite{vicuna}, numerous research works have successfully explored methods to extend LLMs to other modalities, including video, audio, and images. This has led to the development of a variety of related MLLM models, as shown in~\cref{tab:mllm}. For instance, Flamingo~\cite{AlayracDLMBHLMM22} employs cross-attention structures for visual-text interaction, while DetGPT~\cite{pi2023detgpt} connects multimodal LLMs with detectors, enabling object detection based on user instructions. BLIP-2~\cite{li2023blip} encodes input images into queries using the Q-Former structure. mPLUG-Owl~\cite{ye2023mplug} and LLaVA~\cite{abs-2304-08485} encode images into image features using image encoders, which are then combined with text embeddings for LLM learning.  LISA~\cite{lai2023lisa} achieves referring segmentation by training LLMs to output mask token embeddings in an end-to-end manner. NExTGPT~\cite{wu2023next} implements multimodal generation by aligning the output of the adapter, which is added after the LLM, with the output text features from the text encoder of various generation models~\cite{LiuCYMLM0P23, RombachBLEO22}.

%



\section{Methodology}


In this work, we present LLMBind, a unified framework for model integration, aiming to handle various modality tasks into a shared LLM. In \cref{sec:LLMBind}, we detail the components of LLMBind and the process of their integration. In \cref{sec:MoE-LoRA}, we discuss our implementation of the LoRA Mixture-of-Experts technique. In \cref{sec:objectives}, we focus on the training objectives and the specific loss functions we use for model optimization. Finally, we describe the dataset crucial for the training of LLMBind in \cref{sec:dataset}.


\subsection{LLMBind}
\label{sec:LLMBind}
\begin{figure*}[t]
\centering
\includegraphics[width=2.0\columnwidth]{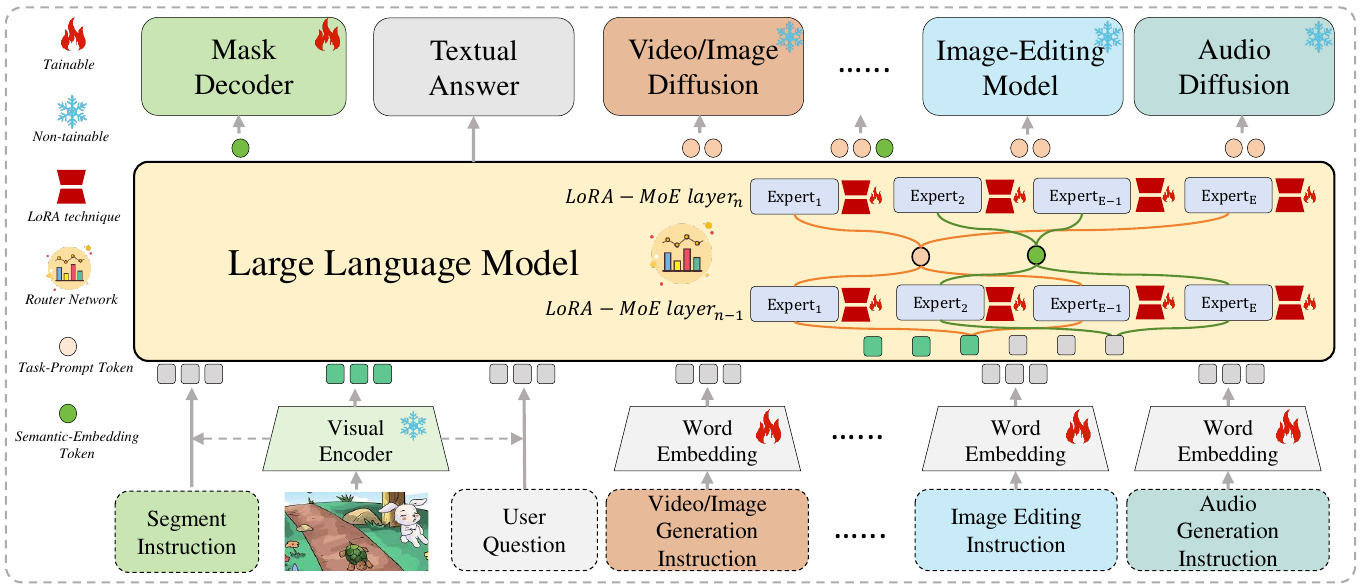}
\caption{
\textbf{LLMBind architecture.} Utilizing task-specific tokens, our framework implements universal multimodal understanding, segmentation, editing, and generation capabilities. During training, we employ LoRA-MoE techniques to facilitate model learning of different modality tasks and reduce computational complexity.
}
\label{fig:framework}

\vskip -0.1in
\end{figure*}

\textbf{Overview.}
The framework of LLMBind is illustrated in~\cref{fig:framework}. It primarily consists of a vision transformer~\cite{radford2021learning} as the image encoder, a large language model called Vicuna~\cite{vicuna}, a SAM-like mask decoder as well as image, audio, and video diffusion models~\cite{RombachBLEO22,abs-2305-11846,abs-2209-14792} for generation and editing tasks.

\noindent \textbf{Visual Encoder and Tokenizer.}
We prioritize the processing of inputs from two primary modalities: images and text. For image encoding, we employ the ViT-L/14~\cite{radford2021learning} to extract image features. We first reshape the image $ x \in \mathbb{R}^{H \times W \times C}$
into a sequence of flattened 2D patches $X_v = [ x_v^1, x_v^2, ..., x_v^N] \in \mathbb{R}^{N \times (P^2 \times C)} $, where $(H, W)$ is the shape of the original image, $C$ is the number of channels, $N = \frac{H \times W}{P^2}$ is the resulting number of image patches.
The LLM uses constant latent vector size $D$ through all of its layers, so we use two MLP layers to map the image patches to $D$ dimensions. Similarly, text is processed through a text embedding function $g$, which projects it into a sequence of text tokens $X_t=[x_t^1,x_t^2,\ldots,x_t^N]\in\mathbb{R}^{N\times D}$, where $N$ is the length of this text token sequence.

\begin{figure}[htbp]
\centering
\includegraphics[width=1\columnwidth]{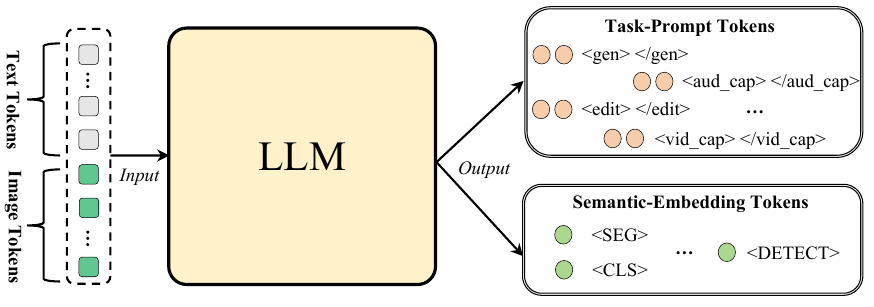}
\caption{ \textbf{Task-Specific tokens} include two categories: Task-Prompt Tokens and Semantic-Embedding Tokens.
\label{fig:task_specific_token}
}

\end{figure}

\noindent \textbf{Task-Specific Tokens.}
Different multimodal tasks often require distinct network structures, making it difficult to consolidate them within a shared LLM. To overcome this challenge, we expand the vocabulary of the LLM with specially designed, learnable task-specific tokens. 
As shown in \cref{fig:task_specific_token}
, these tokens can be classified into two categories: \redtoken{Task-Prompt Tokens} and \greentoken{Semantic-Embedding Tokens}. The first category includes tokens similar to standard text tokens, such as \texttt{\textcolor{redbrown}{<gen>}} and \texttt{\textcolor{redbrown}{</gen>}} for image generation, \texttt{\textcolor{redbrown}{<vid\_cap>}} and \texttt{\textcolor{redbrown}{</vid\_cap>}} for video generation, and \texttt{\textcolor{redbrown}{<edit>}} and \texttt{\textcolor{redbrown}{</edit>}} for image editing. These tokens appear in pairs, and the content between them represents the intermediate text prompt required for the corresponding modality task. 
In contrast, the second category is imbued with dense semantic information, diverging from traditional text tokens.  For example, \texttt{\textcolor{green}{<seg>}} contains rich visual information related to segmented objects in an image. After decoding by a mask decoder, this token can be converted back into masks.  This category also encompasses tokens like \texttt{\textcolor{green}{<detect>}} and \texttt{\textcolor{green}{<cls>}}, designed for extracting spatial coordinates and classification data in image-based tasks. This dual-token framework not only enhances the adaptability of the LLM across various modalities but also ensures scalability and extensibility.

\noindent \textbf{Modality Task Integration.} 
We seamlessly integrate various modality-task models in our framework using task-specific text prompts and token embeddings generated by LLM. For understanding tasks, we directly output text. In the generation and editing scenarios, we feed text prompts into the corresponding task model. Specifically, we use Stable Diffusion~\cite{RombachBLEO22} for image generation, CoDi~\cite{abs-2305-11846} for audio generation, ModelScopeT2V~\cite{wang2023modelscope} for video generation, Instruct-pix2pix~\cite{brooks2023instructpix2pix} for image editing, thus ensuring high-quality, modality-specific output. For complex tasks such as referring segmentation, we use a single task-specific token embedding to learn the semantic information of the image features with respect to the segmentation instructions, which is then processed by a dedicated segmentation decoder. This Integration method is adaptable to various task requirements and has good extensibility.
\subsection{LoRA Mixture-of-Experts}
\label{sec:MoE-LoRA}
To enable effective learning for different multimodal tasks through collaboration among diverse experts, we introduce a LoRA Mixture-of-Experts (MoE) technique. In the traditional transformer architecture, for a given input $x$, the forward propagation process of each decoder block can be written as follows:
\begin{equation}
    f(x) = x+f_{FFN}(x),
\end{equation}
where $f_{FFN}$ denotes the feed-forward network. The matrix operation of the linear layer in the FFN block can be expressed as:
\begin{equation}
    o = Wx = W_0x + \triangle W x, 
\end{equation}
where $W_0 \in \mathbb{R}^{d_{in}, d_{out}}$ denotes the parameter metrix, $\triangle W \in \mathbb{R}^{d_{in}, d_{out}}$ represents the parameter update and $o$ is the output of the linear layer.
After we substitute the linear layer with the MoE technique, the forward process of the MoE layer can be expressed as follows:
\begin{equation}
    o =  W_0x + \triangle W x = W_0 x + \sum_{i=1}^{N} G(x)_i E_i(x),
\end{equation}
\begin{equation}
    G(x) = Softmax(x\cdot W_g),
\end{equation}
where $E_i(\cdot)$ denotes the $i$-th expert, $G(\cdot)$ represents the router network in the MoE layer and $W_g$ is a trainable parameter. Using the router network, these parallel experts can efficiently handle different multimodal tasks through collaboration. To save training cost, following LoRA-MoE~\cite{dou2023loramoe}, we adapt the experts in the MoE layer into a low-rank format and the matrix $\triangle W_{E} \in \mathbb{R}^{d_{in} \times d_{out}}$ of a single expert can be expressed as follows:
\begin{equation}
    \triangle W_{E} = BA,
\end{equation}
where $A \in \mathbb{R}^{d_{in} \times r}$, $ B \in r \times \mathbb{R}^{d_{in}}$ and the rank $r << min(d_{in}, d_{out})$. The forward process of the second modified linear layer can be written as follows:
\begin{equation}
     o = W_0 x + \gamma \sum_{i=1}^{N} G(x)_i E_i(x) = W_0 x +  \frac{\alpha}{r} \sum_{i=1}^{N}{w_i}B_iA_ix,
\end{equation}
where $w_i$ is the attention weight of $i$-th expert and $\alpha$ is a constant hyper-parameter. This low-rank format Mixture-of-Experts greatly reduces the computational overhead imposed by MoE, resulting in a reduction of training parameters to approximately 0.3\% of the total parameter count.

\subsection{Training Objective} 
\label{sec:objectives}
The training of our model involves three objectives: text auto-regressive loss $\mathcal{L}_{reg}$, segmentation mask loss $\mathcal{L}_{mask}$ and MoE auxiliary loss $\mathcal{L}_{aux}$. 
These objectives are combined using weighted coefficients ($\lambda_{reg}$, $\lambda_{mask}$ and $\lambda_{aux}$) to compute overall objective $\mathcal{L}$:

\begin{equation}
\mathcal{L} = \lambda_{reg} \mathcal{L}_{reg} + \lambda_{mask} \mathcal{L}_{mask} + \lambda_{aux} \mathcal{L}_{aux}.
\end{equation}



\textbf{Auto-Regressive Loss.} 
For task-specific text prompt training, we optimize the output of LLM through a generative loss in an auto-regressive manner. Given input tokens $X$,  we compute the auto-regressive loss by:

\begin{equation}
\mathcal{L}_{\text{reg}}(\theta) = \left\{
        \begin{array}{lr}
           -\sum_{i=1}^N log(p_{\theta}(x_i | X_{v}, X_{t, < i})), & \text{if} \  X_v \subseteq X  \\ \\
           -\sum_{i=1}^N log(p_{\theta}(x_i | X_{t, < i})), & \text{if} \  X_v \nsubseteq X  \\
        \end{array}
\right.
\end{equation}

\noindent where $\theta$ is the trainable parameter, $X_{v}$ are the image tokens, $X_{t,<i}$ are the instruction and answer tokens in all turns before the current prediction token $x_i$.


\textbf{Segmentation Loss.} To compute the segmentation loss $\mathcal{L}_{mask}$, a combination of per-pixel Binary Cross-Entropy loss $\mathcal{L}_{BCE}$ and Die Coefficient loss $\mathcal{L}_{DICE}$ is used, with corresponding loss weights, $\lambda_{bce}$ and $\lambda_{dice}$. Given the ground-truth targets $M$ ,
the loss can be expressed as follows:
\begin{equation}
\mathcal{L}_{mask} = \lambda_{bce} \mathcal{L}_{BCE}(\hat{{M}}, {M}) + \lambda_{dice}\mathcal{L}_{DICE}(\hat{{M}}, {M}),
\end{equation}
where  $\hat{{M}}$ represents the predicted segmentation masks. 

\textbf{MoE Auxiliary Loss.} To ensure a balanced load across experts, we integrate an auxiliary loss~~\cite{fedus2022switch}. For each MoE layer, this auxiliary loss is added to the total model loss during training.
With $N$ experts numbered from $i=1$ to $N$, and considering a batch $\mathcal{B}$ consisting of $T$ tokens, we determine this auxiliary loss through the scaled dot-product of two vectors, $f$ and $P$, as follows:
\begin{equation}  \label{eq:total_loss}
\mathcal{L}_{aux} = \alpha \cdot N \cdot \sum_{i=1}^{N} f_i \cdot P_i,
\end{equation}
where $\alpha$ is a multiplicative coefficient, $f_i$ represents the proportion of tokens assigned to expert $i$, calculated by:

\begin{equation} \label{eq:token_sum}
f_i = \frac{1}{T}\sum_{x \in \mathcal{B}} \mathbbm{1} \{\text{argmax}\: p(x) = i\}.
\end{equation}
Meanwhile, $P_i$ indicates the fraction of the router probability allocated for expert $i$.

\begin{equation} \label{eq:prob_sum}
P_i = \frac{1}{T}\sum_{x \in \mathcal{B}} p_i(x).
\end{equation}
Here, $p_i(x)$ is the probability of routing token $x$ to expert $i$. 


\subsection{Dataset}
\label{sec:dataset}
\begin{figure}[t]
\centering
\includegraphics[width=0.7\columnwidth]{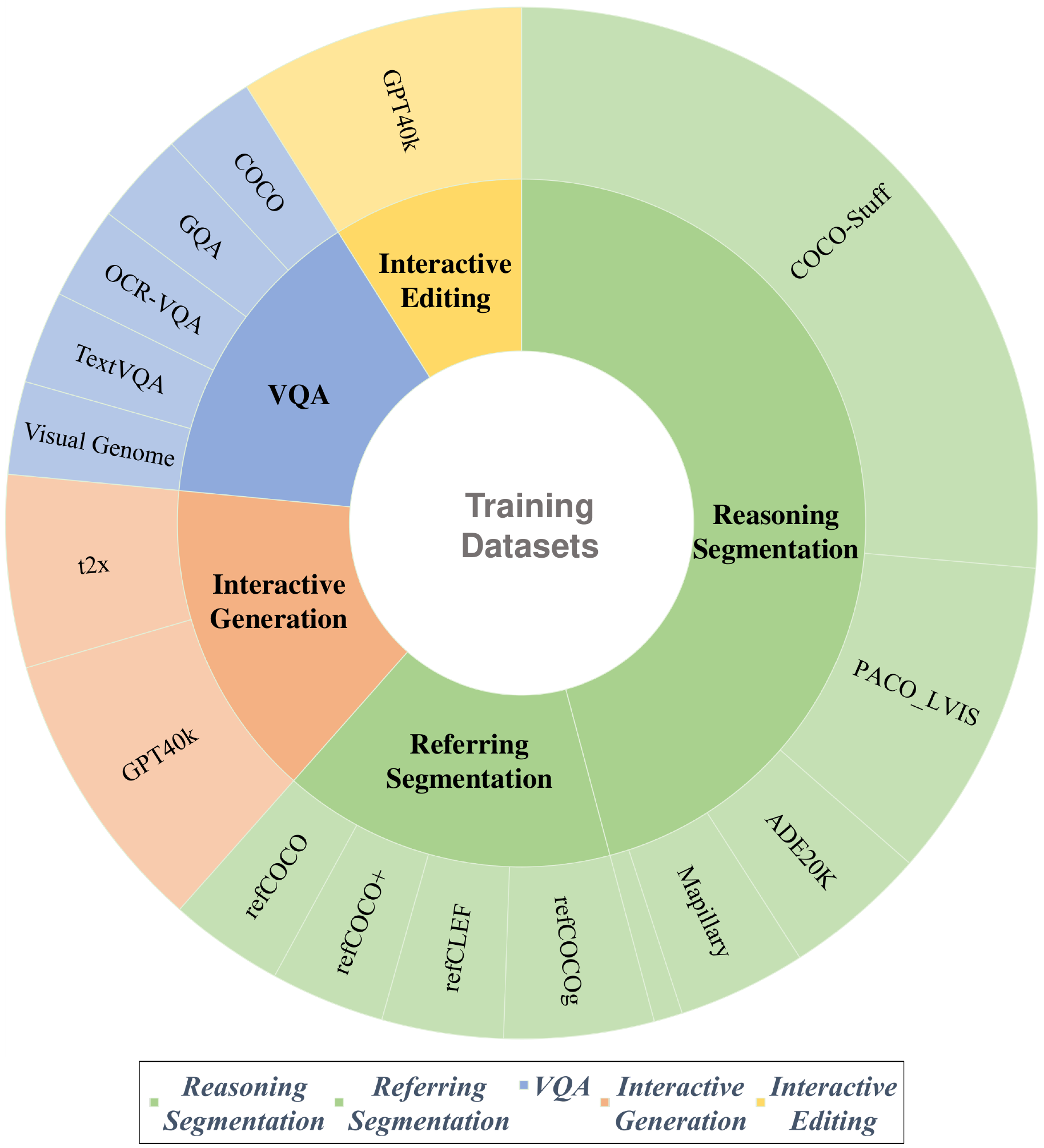}
\caption{ \textbf{Different modality task datasets} used in LLMBind, including VQA, Interactive Generation (video, audio, and image), Interactive Image Editing, and Segmentation.
}
\label{fig:training_data}
\end{figure}

The training datasets consist of both publicly available datasets and our custom proprietary ones, as visualized in ~\cref{fig:training_data}. These proprietary datasets are specifically constructed through interactive processes enabled by ChatGPT.

\begin{figure}[t]
\centering
\includegraphics[width=1.0\columnwidth]{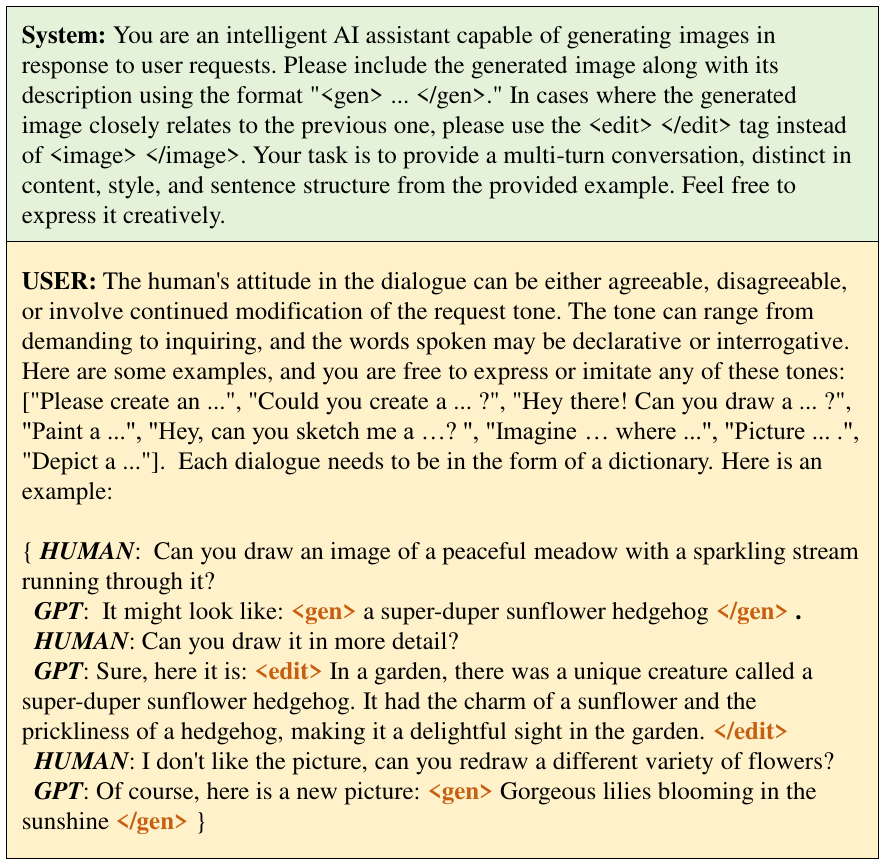}
\caption{\textbf{Prompts for ChatGPT} to construct interactive image generation and editing data.}
\label{fig:prompt_for_chatgpt}
\vskip -0.1in
\end{figure}

\noindent \textbf{Interactive Image Generation and Editing Dataset.}
To empower the model in executing image generation and editing tasks, following Mini-DALLE3~\cite{zeqiang2023mini}, we design a prompt for ChatGPT to construct 400k human-like interactive instruction data, tackling the problem of image generation and editing through textual description. 
As shown in ~\cref{fig:prompt_for_chatgpt}, we first clarify the role and capability of ChatGPT within the prompt. We provide it with instruction text formats, tags, and examples for both generating and editing tasks. However, we notice that the generated samples closely resemble the examples in terms of text format, tone, and content. To address this, we further clarify the diversity of the target data in terms of the above multiple aspects in the prompt. Following a manual review of a portion of our dataset, we have confirmed that this approach is highly effective in ensuring significant diversity.

\noindent \textbf{Interactive Video and Audio Generation Dataset.} To construct video and audio generation instruction data, we adapt the t2x\cite{wu2023next} dataset and change its format to resemble that used for the image generation task. Some examples are listed in Appendix C.




\noindent \textbf{Segmentation Datasets.} We incorporate several semantic segmentation datasets including ADE20K~\citep{zhou2017scene}, COCO-Stuff~\citep{caesar2018coco}, PACO-LVIS~\citep{ramanathan2023paco}, PartImageNet~\citep{he2022partimagenet}, and PASCAL-Part~\citep{chen2014detect}
and utilize referring segmentation datasets like refCOCO, refCOCO+\citep{kazemzadeh2014referitgame}, and refCOCOg\citep{mao2016generation}. 



\section{Experiments}

\begin{table*}[t!]

\setlength\tabcolsep{5mm}
\caption{\textbf{Referring segmentation results} on three referring segmentation datasets: refCOCO \cite{kazemzadeh2014referitgame}, refCOCO+ \cite{kazemzadeh2014referitgame}, and refCOCOg \cite{mao2016generation} with metric  cIoU. The best
results and second best results are indicated by \textbf{boldface} and \underline{underline}, respectively.}

\begin{center}
\vskip -0.1in
    {
        \begin{adjustbox}{scale=0.85}
        \begin{tabular}{ l | c c c | c c c | c c }
            \toprule
            \multirow{3}*{\textbf{Method}} & \multicolumn{3}{c|}{\textbf{refCOCO}} & \multicolumn{3}{c|}{\textbf{refCOCO+}}  & \multicolumn{2}{c}{\textbf{refCOCOg}} \\ 
            
            \specialrule{0em}{0pt}{1pt}
            \cline{2-9}
            \specialrule{0em}{0pt}{1pt}
            
            ~ & val & testA & testB & val & testA & testB & val(U) & test(U) \\

            \specialrule{0em}{0pt}{1pt}
            \hline
            \specialrule{0em}{0pt}{1pt}

            MCN~\citep{luo2020multi} & 62.4 & 64.2 & 59.7 & 50.6 & 55.0 & 44.7 & 49.2 & 49.4 \\

            VLT~\citep{ding2021vision} & 67.5 & 70.5 & 65.2 & 56.3 & 61.0 & 50.1 & 55.0 & 57.7 \\

            CRIS~\citep{wang2022cris} & 70.5 & 73.2 & 66.1 & 62.3 & 68.1 & 53.7 & 59.9 & 60.4 \\

            LAVT~\citep{yang2022lavt} & 72.7 & 75.8 & 68.8 & 62.1 & 68.4 & 55.1 & 61.2 & 62.1 \\
            

            ReLA~\citep{liu2023gres} & 73.8 & 76.5 & 70.2 & 66.0 & 71.0 & 57.7 & 65.0 &  66.0  \\
            
            X-Decoder~\citep{zou2023generalized} & - & - & - & - & - & - & 64.6 & -  \\

            SEEM~\citep{zou2023segment} & - & - & - & - & - & - & 65.7 & -    \\
            

             LISA~\cite{lai2023lisa} &  74.9   &  \textbf{79.1}  &  72.3 & 65.1 & 70.8 & 58.1 & 67.9  &  \underline{70.6}  \\

            UnifiedMLLM~\cite{li2025unifiedmllm} &  \underline{76.3}  &  \underline{78.8}  &  \underline{72.7}  &  \underline{66.4} & \textbf{72.4} & \underline{59.1} &  \underline{68.0}  &  69.6 \\

            \specialrule{0em}{0pt}{1pt}
            \hline
            \specialrule{0em}{0pt}{1pt}

           \rowcolor{aliceblue} LLMBind (ours) & \textbf{76.9}  &   78.5   & \textbf{73.2} & \textbf{67.8} & \underline{71.9} & \textbf{60.8}   & \textbf{69.8}  &   \textbf{70.8} \\
            
            \bottomrule            
        \end{tabular}
        \end{adjustbox}
    }
    \label{table:reason_seg}   
\end{center}
\vskip -0.1in
\end{table*}

\subsection{Experimental Setting}
\label{exp:setting}
\textbf{Training Specifications.} During training, we employ 8 NVIDIA A100 GPUs (40G). The training scripts are based on the deepspeed~\citep{rasley2020deepspeed} engine. We use the AdamW~\citep{loshchilov2017decoupled} optimizer with a learning rate of 0.0003 and weight decay of 0. The WarmupDecayLR learning rate scheduler is adopted, with a warmup iteration count of 100. The auto-regressive loss and segmentation loss have weights of 1.0 each. The BCE loss and DICE loss have weights of 2.0 and 0.5 respectively. We set the batch size per device to 2, and accumulate gradients over 10 steps, with a total of 15k steps.

\noindent \textbf{Data Details.}
During training, to sustain the stability of multi-task training, we configure the data ratios for Segmentation data, Visual Question Answering data, and Interactive Generation and Editing instruction data at a 1:1:1 ratio. Within the segmentation dataset, the proportions for semantic segmentation, referring segmentation, and reasoning segmentation are established at 9:3:1. In the interactive instructions, the distribution for tasks like image generation, image editing, audio generation, and video generation is balanced at a 1:1:1:1 ratio. We train jointly on all task datasets to ensure that our model avoids loss fluctuations and balances losses across tasks.

\noindent \textbf{Evaluation Metrics.} 
Text-to-audio generation is assessed using Fréchet Distance (FD) for distribution similarity and Inception Score (IS) for quality and diversity. Text-to-video generation leverages Fréchet Inception Distance (FID) for content quality and CLIPSIM for textual alignment. In text-to-image generation, FID evaluates image distribution similarity. For segmentation, gIoU is defined by the average of all per-image Intersection-over-Unions (IoUs), while cIoU is defined by the cumulative intersection over the cumulative union. In addition, to better showcase the model's generative capabilities during user interaction, we conducted a human evaluation to assess the quality of the model output and its alignment with the given text instructions.

\begin{table}[t]
\centering
\caption{\textbf{Text-to-image generation results} on COCO-captions dataset, evaluated with the FID for image quality.}

\setlength\tabcolsep{5mm}
\label{table:text_to_image}

\begin{adjustbox}{scale=0.85}
\begin{tabular}{lc}
\hline
\bf Method & \bf FID ($\downarrow$) \\
\hline
\multicolumn{2}{l}{\textit{Specialized Models}} \\
CogVideo~\cite{DingYHZZYLZSYT21} & 27.10 \\   
GLIDE~\cite{NicholDRSMMSC22} & 12.24 \\
CoDi~\cite{abs-2305-11846} & 11.26 \\
\hline
\multicolumn{2}{l}{\textit{Multitask Models}} \\
NExT-GPT~\cite{wu2023next} & 11.28 \\
UnifiedMLLM~\cite{li2025unifiedmllm}   & 10.84 \\
\rowcolor{aliceblue} LLMBind-GALIP* & \textbf{10.38}  \\
\hline
\end{tabular}
\end{adjustbox}

\vskip -0.1in
\end{table}


\subsection{Quantitative Analysis} 
\textbf{Referring Segmentation.}
As illustrated in \cref{table:reason_seg}, on the refCOCO dataset, for the val, testA, and testB splits, LLMBind achieves scores of 76.9, 78.5, and 73.2, respectively. Similarly, on the refCOCO dataset, the model records scores of 67.8, 71.9, and 60.8 for val, testA, and testB. On the refCOCOG dataset, LLMBind achieves 69.8 and 70.8.

\noindent \textbf{Image Generation.} 
As shown in~\cref{table:text_to_image}, using GALIP~\cite{tao2023galip} as the baseline model, LLMBind secures an FID score of 10.38, outperforming models such as NeXT-GPT~\cite{wu2023next} and CoDi~\cite{abs-2305-11846}.

\begin{table}[t]
    
    \centering
    \setlength\tabcolsep{5mm}
    \label{table:text_to_audio}
    \caption{\textbf{Text-to-audio generation results} on AudioCaps dataset, evaluated with FD for audio similarity and IS for quality.}
    \begin{adjustbox}{scale=0.85}
    \begin{tabular}{lcc}
    \hline
    \bf Method & \bf FD ($\downarrow$) & \bf IS ($\uparrow$) \\
    \hline

    \multicolumn{2}{l}{\textit{Specialized Models}} \\

    DiffSound~\cite{YangYWWWZY23}&	47.68 & 4.01 \\   
    AudioLDM-S~\cite{LiuCYMLM0P23} & 29.48 & 6.90 \\   
    AudioLDM-L~\cite{LiuCYMLM0P23} & {23.31}  &  8.13 \\   
    \hline
    \multicolumn{2}{l}{\textit{Multitask Models}} \\
    NExT-GPT~\cite{wu2023next} & 23.58 & {8.35} \\   
    UnifiedMLLM~\cite{li2025unifiedmllm}  &  22.42 &  9.95  \\
    \rowcolor{aliceblue} LLMBind-CoDi & 22.90 & 8.77 \\
    
    \rowcolor{aliceblue} LLMBind-Auffusion & \bf 21.99 & \bf 10.57 \\
    \hline
    \end{tabular}
    \end{adjustbox}
 \vskip -0.1in
\end{table}

\noindent \textbf{Audio Generation.} 
We evaluate LLMBind on AudioCaps~\cite{KimKLK19} using different generative backbones. As shown in~\cref{table:text_to_audio}, with CoDi~\cite{abs-2305-11846} as the baseline, LLMBind achieves an FD of 22.90 and an IS of 8.77. By switching to Auffusion~\cite{abs-2305-11846}, performance further improves to 21.99 (FD) and 10.57 (IS), consistently outperforming specialized models like AudioLDM~\cite{LiuCYMLM0P23} and unified frameworks such as NExT-GPT~\cite{wu2023next}.

\begin{table}[t]
\centering

\caption{\textbf{Text-to-video generation results} on MSR-VTT, evaluated with FID for video quality and CLIPSIM for video coherence.}
\setlength{\tabcolsep}{2mm}
\label{table:text_to_video}

\begin{adjustbox}{scale=0.85}
\begin{tabular}{lcc}
\hline
\bf Method & \bf FID ($\downarrow$) & \bf  CLIPSIM ($\uparrow$) \\
\hline

\multicolumn{2}{l}{\textit{Specialized Models}} \\
CogVideo~\cite{abs-2205-15868}&	23.59 & 	0.2631 \\   
   
Latent-VDM~\cite{RombachBLEO22} &  14.25 & 	0.2756 \\   
Latent-Shift~\cite{abs-2304-08477} & 15.23 & 	0.2773 \\
Make-A-Video~\cite{singer2022make} & 13.17 & 0.3049\\
CoDi~\cite{abs-2305-11846} & - & 0.2890 \\
\hline
\multicolumn{2}{l}{\textit{Multitask Models}} \\
NExT-GPT~\cite{wu2023next} & 13.04 &  \bf{0.3085} \\    
UnifiedMLLM~\cite{li2025unifiedmllm}  &  11.15 & 0.312 \\
\rowcolor{aliceblue} LLMBind-ModelScopeT2V & 	 11.09 &   0.2930 \\
\rowcolor{aliceblue} LLMBind-HiGen & 	\bf 8.60 &  0.2947  \\
\hline
\end{tabular}
\end{adjustbox}
\end{table}
\noindent \textbf{Video Generation.} 
We report LLMBind's performance on MSR-VTT~\cite{XuMYR16} across various generative backbones in Table~\ref{table:text_to_video}. With ModelScopeT2V~\cite{wang2023modelscope}, LLMBind achieves an FID of 11.09 and CLIPSIM of 0.2930. Switching to HiGen~\cite{qing2023hierarchical}, the performance further improves to a superior FID of 8.60, outperforming NExT-GPT~\cite{wu2023next}.


\begin{figure}
\centering
\includegraphics[width=1.0\columnwidth]{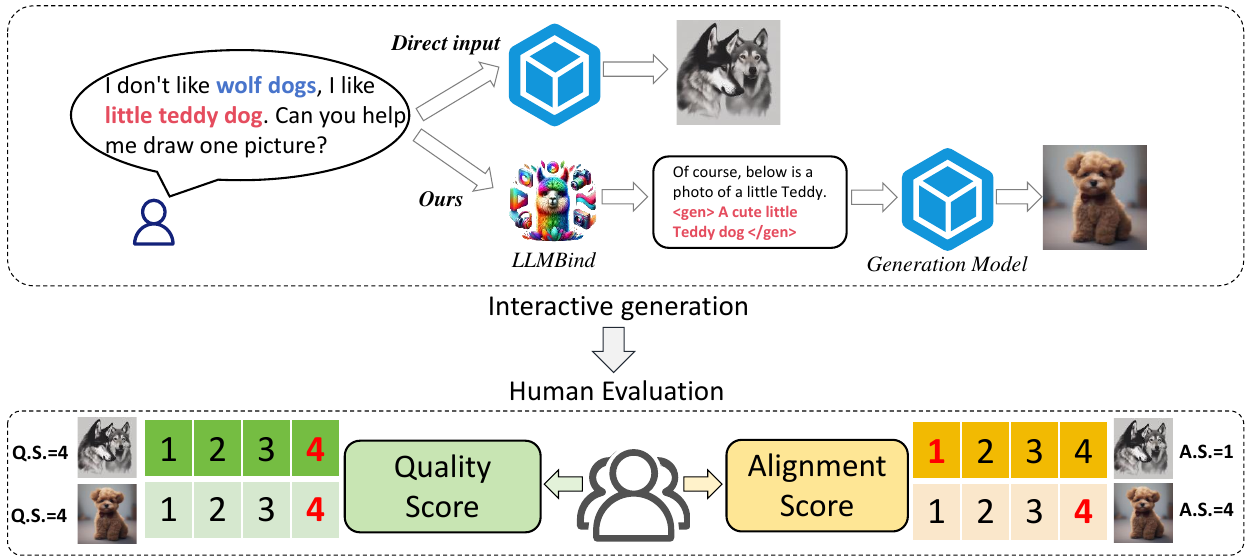}
\caption{ \textbf{Human evaluation.} The evaluation indicators for interactive generation include Q.S. and A.S.
}
\label{fig:human evaluation}
\vskip -0.1in
\end{figure}

\subsection{Human Evaluation}

\begin{table*}[t!]
\begin{minipage}{\textwidth}
\vskip 0.05in
\begin{minipage}[t]{0.3\textwidth}
\centering
\fontsize{7}{11}\selectfont
\setlength{\tabcolsep}{0.5mm}
\caption{\textbf{Ablation on LoRA rank.}}
\vskip -0.05in
\label{table:rank_of_lora}
\begin{tabular}{  >{\columncolor{gray!10}}c c c  c c | c c}
    \toprule
     LoRA &  Training & \multirow{2}{*}  \textbf{MoE}  & Expert & Top-K  &   \multirow{2}{*}{ gIoU } &   \multirow{2}{*}{cIoU}  \\
     RANK &   Steps &   Layers &  Nums  &  Experts \\
    \midrule

      4 & 7k  & 16/32 & 4 &2  & 59.4 & 62.5  \\
    
    \rowcolor{aliceblue} \textbf{8} & 7k  & 16/32  & 4 & 2 &  \textbf{58.9} & 65.8 \\

    12 & 7k   & 16/32  & 4 &2   & 58.7 & \textbf{66.2} \\
    
   16 & 7k   & 16/32  & 4 &2    & 58.6 & 65.2  \\

    \bottomrule
\end{tabular}
\end{minipage}
\hspace{13pt}
\begin{minipage}[t]{0.3\textwidth}
\centering
\fontsize{7}{11}\selectfont
\setlength{\tabcolsep}{0.5mm}



    

\caption{\textbf{Experts and Top-K routing.}}
\vskip -0.05in
\label{table:expert_nums_and_topk}
\begin{tabular}{  c c c  >{\columncolor{gray!10}}c >{\columncolor{gray!10}}c | c c}
    \toprule
      LoRA &  Training & \multirow{2}{*}  \textbf{MoE}  & Expert & Top-K  &   \multirow{2}{*}{ gIoU } &   \multirow{2}{*}{cIoU}  \\
     RANK &   Steps &   Layers &  Nums  &  Experts \\
    \midrule

    \rowcolor{aliceblue} 8 & 7k  & 16/32 & \textbf{4} & \textbf{2}  & \textbf{59.4} & \textbf{66.5}  \\
    
    8 & 7k  & 16/32  & 3 & 2 &  58.9 & 65.8 \\

    8 & 7k   & 16/32  & 2 &2   & 58.7 & 66.2 \\
    
   8 & 7k   & 16/32  & 2 &1    & 58.6 & 65.2  \\

    \bottomrule
\end{tabular}

\end{minipage}
\hspace{13pt}
\begin{minipage}[t]{0.3\textwidth}
\centering
\caption{\textbf{Ablation on MoE layers.}}
\vskip -0.05in
\label{table:moe_layers}
\fontsize{7}{11}\selectfont
\setlength{\tabcolsep}{0.5mm}
\begin{tabular}{   c c >{\columncolor{gray!10}}c c c | c c}
    \toprule
    LoRA &  Training & \multirow{2}{*}  \textbf{MoE}  & Expert & Top-K  &   \multirow{2}{*}{ gIoU } &   \multirow{2}{*}{cIoU}  \\
     RANK &   Steps &   Layers &  Nums  &  Experts \\
    \midrule

    \rowcolor{aliceblue}  8 & 7k &  \textbf{16/32} & 4 &2  &\textbf{59.9} & \textbf{65.8} \\
    8 & 7k & 10/32 & 4 &2  & 59.2 & 60.9 \\
    8 & 7k &  8/32 & 4 &2   & 58.9 & 57.1  \\
    8 & 7k & 6/32 & 4 &2  &  58.3 & 61.8  \\
    \bottomrule
\end{tabular}
\end{minipage}
\end{minipage}

\vskip -0.1in
\end{table*}




    

    


As shown in ~\cref{fig:human evaluation}, to assess the practical performance of LLMBind in interactive \textbf{I}mage, \textbf{V}ideo, and \textbf{A}udio generation tasks (\textbf{IVA}), we conducted a comprehensive user study. This study is designed to measure the capability of models to generate IVA data in response to interactive text instructions, emulating natural human-like communication. The study focuses on assessing two critical metrics:  Quality Score (Q.S.) and Alignment Score (A.S.). The Q.S. assesses the quality of the generated IVA content, while A.S. evaluates the alignment between the generated data and the textual instructions provided. Participants interact with the model by issuing text instructions and then rate each generated sample on a scale of 1 to 4 for both metrics, with 1 representing poor performance and 4 indicating excellent performance. The aggregated results from a panel of 30 participants, as illustrated in ~\cref{table:human_eval}, clearly indicate that LLMBind excels in both the quality of modal data generation and its ability to align with text-based instructions closely. The final scores, calculated as the weighted average of the ratings for Q.S. and A.S., demonstrate LLMBind's proficiency in understanding and executing complex, multimodal generation tasks as directed by human users.

\begin{table}
\begin{center}
\setlength\tabcolsep{0.9mm}
\caption{\textbf{Human evaluation statistics} regarding the preference rate for Quality Score (Q.S.) and Alignment Score (A.S.). The former metric assesses the quality of generated content, while the latter evaluates the alignment between the generated data and textual instructions. Both metrics are rated on a scale of 1-4, with the final score being the weighted average of these ratings.}
\small
    {

        \begin{adjustbox}{scale=0.85}

        \begin{tabular}{ l | c c c c | c | c c c c | c |}
        \toprule
        \multirow{2}{*}{\textbf{Method}} & \multicolumn{4}{c|}{\textbf{Q.S.}} &\bf Final & \multicolumn{4}{c|}{\textbf{A.S.}} & \bf Final \\
        ~ & \textbf{1} & \textbf{2} & \textbf{3} & \textbf{4} & \bf score &  \textbf{1} & \textbf{2} & \textbf{3} & \textbf{4} & \bf Score\\
        \hline
        \multicolumn{11}{l}{\textit{Audio Generation}} \\
        Direct input               & 7\% & 35\%  & 34\% & 24\% & 2.75 & 18\% & 39\% & 23\% & 20\% & 2.45  \\   
        \rowcolor{aliceblue} Ours  & 8\%  &  28\%  & 35\% & 29\%  & \bf 2.85 & 6\% & 24\% & 27\% & 43\% & \bf 3.07 \\  
        \midrule
        \multicolumn{11}{l}{\textit{Image Generation}} \\
        Direct input               & 5\% & 39\%  & 43\% & 13\% & 2.64 & 21\% & 29\% & 45\% & 5\% & 2.34  \\
        \rowcolor{aliceblue} Ours  & 5\%  &  27\%  & 41\% & 27\%  & \bf 2.90 & 12\% & 19\% & 51\% & 18\% & \bf 2.75 \\  
        \midrule
        \multicolumn{11}{l}{\textit{Video Generation}} \\
        Direct input                 & 29\% & 41\%  & 23\% & 7\% & 2.08 & 23\% & 42\% & 21\% & 14\% & 2.26  \\
        \rowcolor{aliceblue} Ours    & 24\%  & 38\%  & 27\% & 11\%  & \bf 2.25 & 19\% & 40\% & 24\% & 17\% & \bf 2.39 \\  

        \bottomrule            
        \end{tabular}

        \end{adjustbox}
    }
        
    \label{table:human_eval}   

\end{center}
\vskip -0.15in
\end{table}

\subsection{Ablation Studies}

We conduct the ablation study specifically on the reasoning segmentation task, as it provides a robust framework for quantitative analysis and allows for direct comparison with existing benchmarks. Specifically, we analyze how the distribution of MoE layers and the configuration of experts influence the model's performance in handling heterogeneous multimodal tasks.

\noindent \textbf{Effect of LoRA Rank.}

We analyze the impact of the LoRA rank on expert capacity. As shown in Table~\ref{table:rank_of_lora}, increasing the rank yields consistent but diminishing performance gains, indicating that a moderate rank is sufficient for effective expert specialization while maintaining parameter efficiency.


\noindent \textbf{Analysis of MoE Layer Distribution.}
As illustrated in Table~\ref{table:moe_layers}, we investigate the impact of the density of MoE layers within the 32-layer LLM backbone. We observe a consistent performance gain as the number of MoE layers increases. Specifically, increasing the MoE layers from 6 to 16 results in a significant improvement in both gIoU (from 58.3 to 59.9) and cIoU (from 61.8 to 65.8). This trend suggests that a higher density of MoE layers provides the network with more opportunities to decouple gradients from diverse tasks. By routing features through specialized experts at multiple stages of the transformer hierarchy, LLMBind effectively prevents the "modality interference" mentioned in our introduction, ensuring that localization-sensitive features for segmentation are not overshadowed by the semantic gradients of generative tasks.

\noindent \textbf{Effect of Expert Numbers and Routing Strategy.}
Table~\ref{table:expert_nums_and_topk} explores the number of experts and the Top-K routing mechanism. When the expert count is increased from 2 to 4, we observe a steady increase in performance, with gIoU rising from 58.7 to 59.4. This confirms that a larger pool of experts provides greater capacity for \textit{modality disentanglement}, allowing the model to allocate dedicated parameters to the specific nuances of pixel-level perception versus waveform-level synthesis. We compare the performance of Top-1 and Top-2 routing strategies. The results indicate that Top-2 routing (gIoU 58.7) outperforms Top-1 routing (gIoU 58.6). This suggests that for complex multimodal instructions, activating multiple experts allows for better collaborative reasoning. For instance, a task requiring both "referring" (localization) and "editing" (generation) benefits from the simultaneous activation of experts specialized in different feature domains. Consistent with our core philosophy, these results demonstrate that the MoE-LoRA architecture is essential for transforming a standard LLM into a robust multimodal system, maintaining high precision across diverse tasks without performance degradation.

\section{Conclusion and Future Direction}

In this work, we present LLMBind, a unified framework that redefines the MLLM as a versatile multimodal operating system. By strategically balancing \greentoken{\textit{In-Situ}} perception through Semantic Embedding Tokens and \redtoken{\textit{Ex-Situ}} generation via Task-Prompt Tokens, LLMBind overcomes the traditional trade-off between localization precision and integration flexibility. To resolve the inherent challenges of task interference in multi-task learning, we introduce a Task-Specific MoE-LoRA architecture, which ensures effective modality disentanglement and preserves specialized knowledge across heterogeneous domains. 

Furthermore, our curated 400k Interactive Multi-turn Dataset provides a foundation for complex, human-like AI collaboration. Extensive experiments, particularly in reasoning segmentation and multimodal synthesis, demonstrate that LLMBind not only outperforms specialized models but also maintains high updateability for emerging modalities. Moving forward, we aim to extend this framework to a broader spectrum of sensory inputs and continue refining joint training strategies to further harmonize the synergy between perception and generation in generalist agents.

\nocite{langley00}

\bibliography{icml2026}
\bibliographystyle{icml2026}

\newpage
\appendix
\onecolumn

\clearpage
\appendix

\def\apptoptitlebar{\hrule height0pt}
\def\appbottomtitlebar{\vskip .22in }

{\center\baselineskip 18pt
   \apptoptitlebar{\Large\textbf{Appendix for LLMBind}}\appbottomtitlebar}
\section{Interactive Architecture}


\subsection{Multi-Turn and Interactive}

LLMBind allows users to accomplish various modality-tasks as mentioned above through a dynamic and interactive conversational approach. Unlike the traditional fixed-format prompt inputs, users have more freedom to express their requirements and can improve the model's outputs through interactive dialogue, such as modifying the content of generated images. Additionally, LLMBind can automatically preprocess the user's input text intent and analyze the corresponding modality-task, greatly reducing the user's learning curve and enhancing the overall user experience.




\subsection{Interactive Multi-modal Generation} LLMBind can automatically determine the corresponding data type for generation based on the details provided in the input text. This includes image generation, video generation, and audio generation. Unlike traditional generation methods that require specific text descriptions, LLMBind leverages the capabilities of LLM to create image captions that align with user-provided objects, scenes, or concepts. This assists users in conceptualizing image content and achieving their desired image generation outcome. For image generation, LLMBind generates textual descriptions that match the user's intent by incorporating \texttt{\textcolor{redbrown}{<gen>}}DETAILED IMAGE CAPTION\texttt{\textcolor{redbrown}{</gen>}} in the output response. Subsequently, LLMBind employs a generation model prompt style using ``DETAILED IMAGE CAPTION'' to generate the corresponding image. It is worth noting that during this process, users are not required to adhere to a fixed prompt format. For video and text generation, LLMBind follows a similar approach, replacing \texttt{\textcolor{redbrown}{<gen>,</gen>}} with \texttt{\textcolor{redbrown}{<vid\_cap>,</vid\_cap>}} and \texttt{\textcolor{redbrown}{<aud\_cap>,</aud\_cap>}} respectively.

\subsection{Interactive Image Editing and Refinement}
LLMBind offers interactive image editing and refinement capabilities. Image editing typically involves modifying image content based on textual instructions. This can involve significant changes to objects and attributes within the image. The edited image may differ greatly from the original, but it closely aligns with the user's textual instructions. On the other hand, image refinement entails making adjustments and optimizations to enhance the details of the image. The edited image should maintain a significant resemblance to the original but incorporates specific image adjustments based on the provided text instructions.

\subsection{Interactive Text Conversation}
Text-based conversation is a potential capability of LLMs, and LLMBind excels in maintaining effective communication while simultaneously supporting other modality tasks. This functionality enables seamless switching between different modality tasks and highlights the user-friendly nature of the model.




\section{Impact Statements}

\subsection{Broader Impacts}

This study explores the development of our LLMBind model in the field of artificial intelligence, particularly its multimodal capabilities. The development and implementation of the model could have significant impacts on multiple areas of society, both positive and potentially negative.

\begin{itemize}
\item  Creative Industries and Intellectual Property: The LLMBind model's capabilities in content generation have far-reaching implications for the creative industries. It can facilitate the democratisation of content creation and the innovation process, but at the same time may challenge existing norms of intellectual property and originality, raising concerns about authors' rights and creative ethics.
\item Information Integrity and the Media: The model's ability to edit and generate realistic content could have implications for the journalism and media sectors. While being able to automate and improve the accuracy of information dissemination, there is also a risk of being used to create misleading or false information, posing a threat to the authenticity of public discourse.
\item Human-AI Interaction and Social Implications: As the LLMBind model advances in dialogue, its integration in everyday life may affect social dynamics and interpersonal interactions. While they can provide companionship and support in therapeutic and educational settings, they also raise questions about privacy, emotional attachment, and reduced interpersonal interactions.
\end{itemize}

\subsection{Reproducibility}

We have ensured the availability of all models and source codes in a public domain. The provided materials include code that enables the replication of our results.

\subsection{Licenses}
The majority of this project is released under the Apache 2.0 license.
\begin{itemize}
\item The service is a research preview intended for non-commercial use only, subject to the model License of LLaMA (\url{https://github.com/facebookresearch/llama/blob/main/MODEL_CARD.md}). 
\item Terms of Use of the data generated by OpenAI (\url{https://openai.com/policies/terms-of-use}). 

\end{itemize}

\onecolumn
\section{Demo and Training Datasets}




\begin{figure}[htbp]  
    \centering  
    \includegraphics[width=1\textwidth]{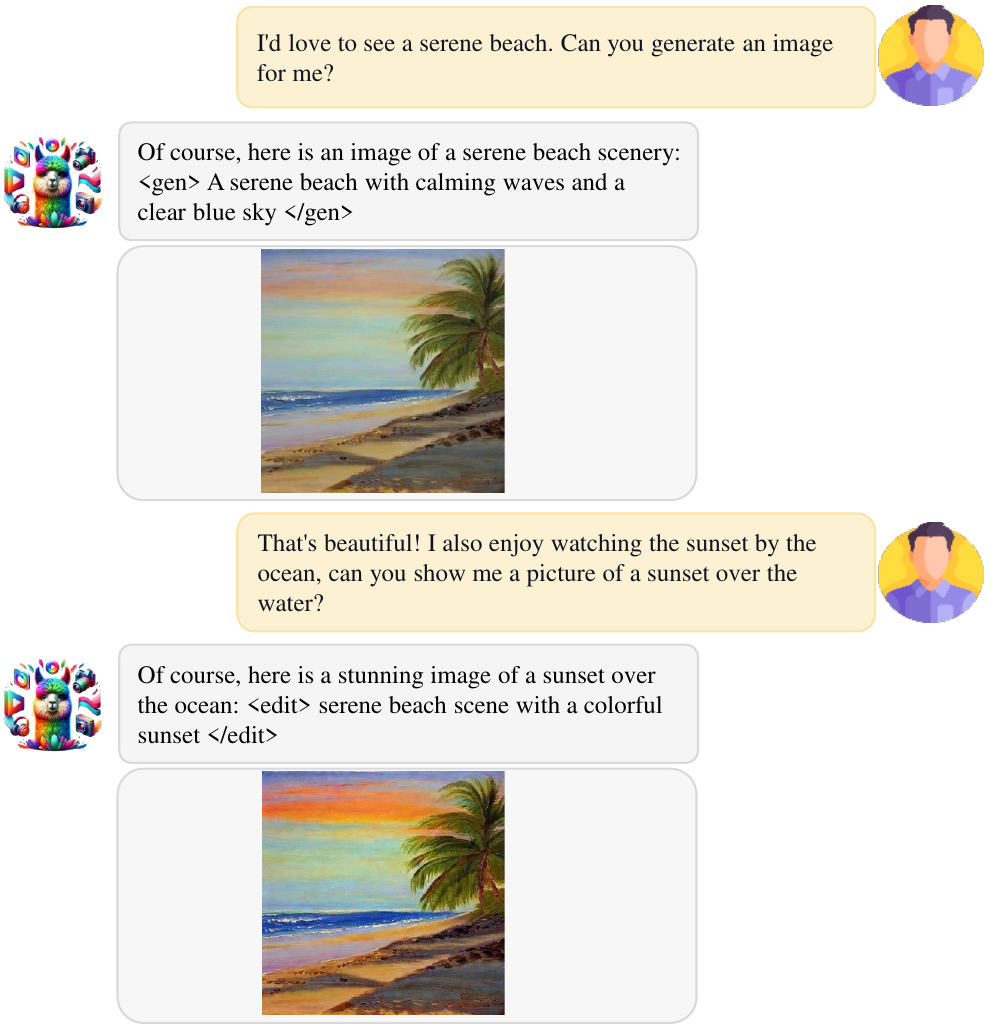}  
    \caption{\textbf{Example of Image Generation and Editing (1).}}  
    \label{fig:gen_edit_sample1}  
\end{figure}  
  
\begin{figure}[htbp]  
    \centering  
    \includegraphics[width=1.0\textwidth]{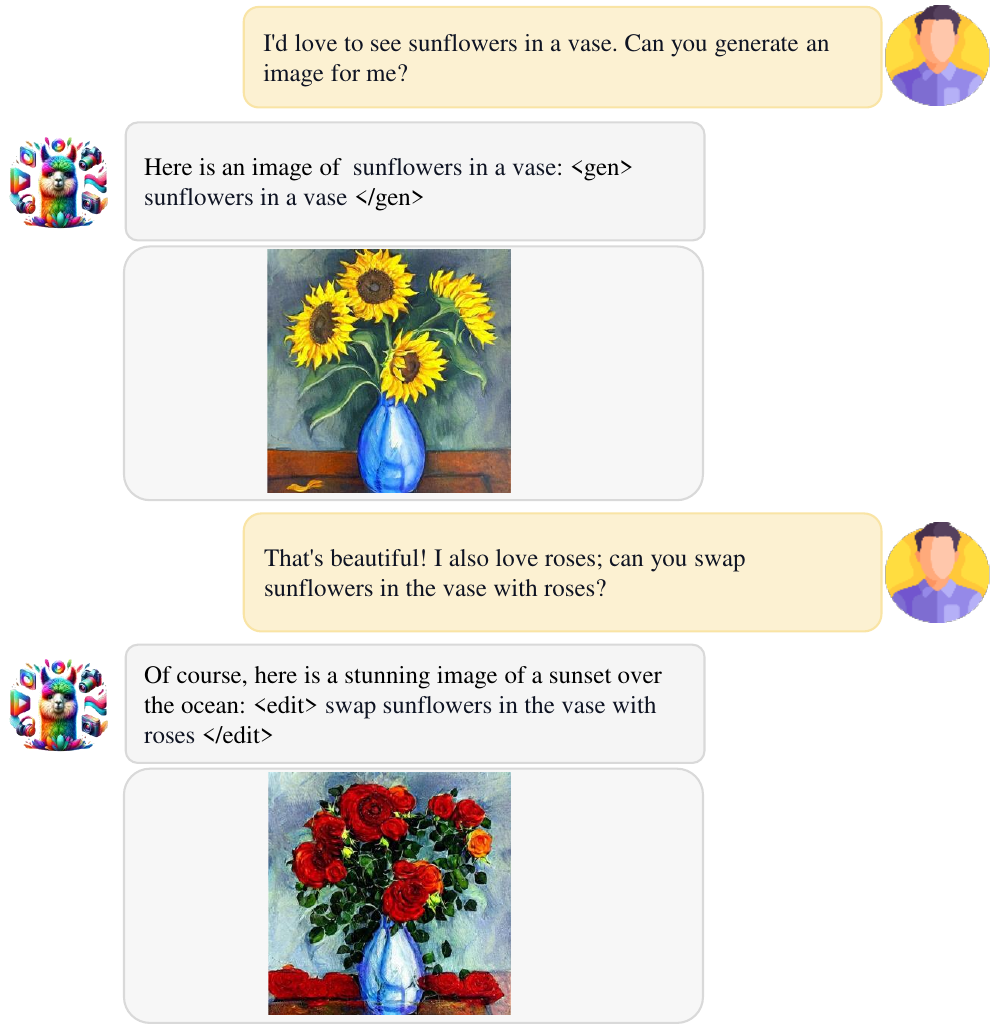}  
    \caption{\textbf{Example of Image Generation and Editing (2).}}  
    \label{fig:gen_edit_sample2}  
\end{figure}  

\begin{figure}[htbp]  
    \centering  
    \includegraphics[width=1\textwidth]{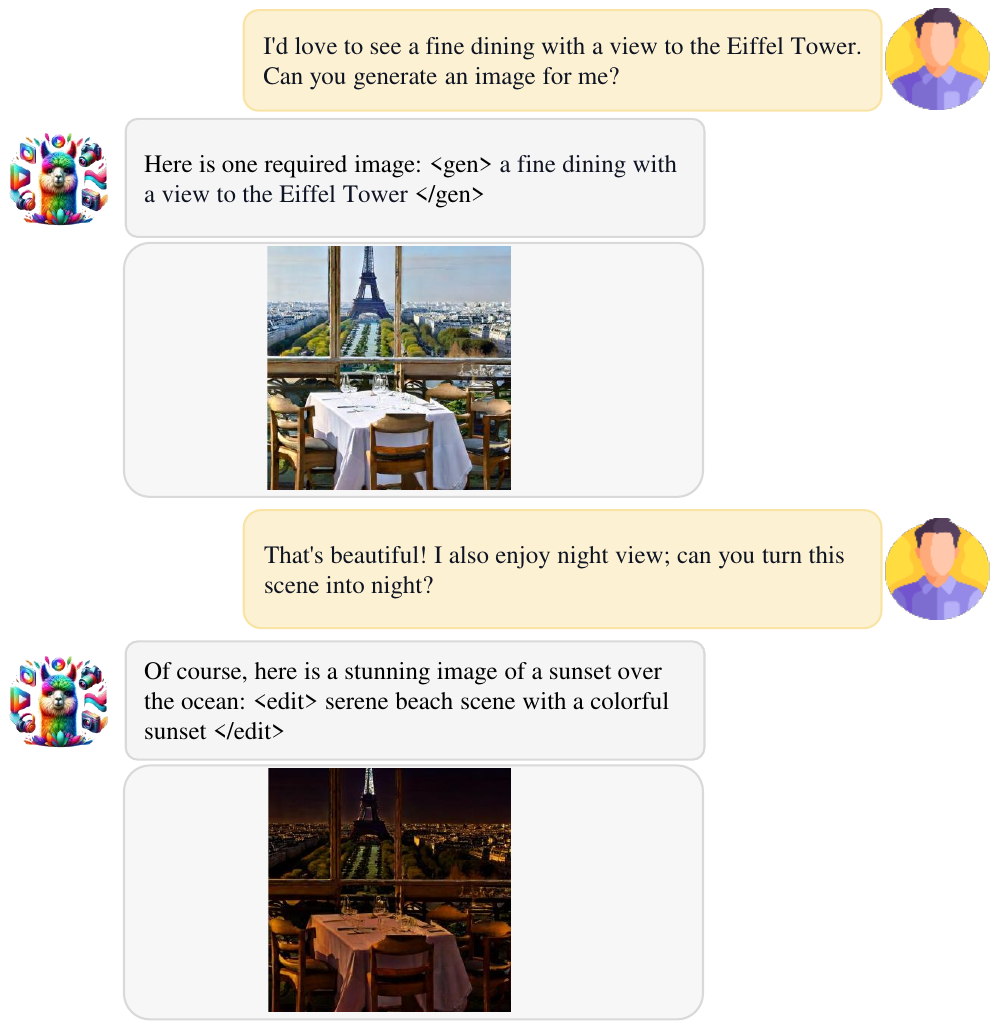}  
    \caption{\textbf{Example of Image Generation and Editing (3).}}  
    \label{fig:gen_edit_sample1}  
\end{figure}  
  
\begin{figure}[htbp]  
    \centering  
    \includegraphics[width=1.0\textwidth]{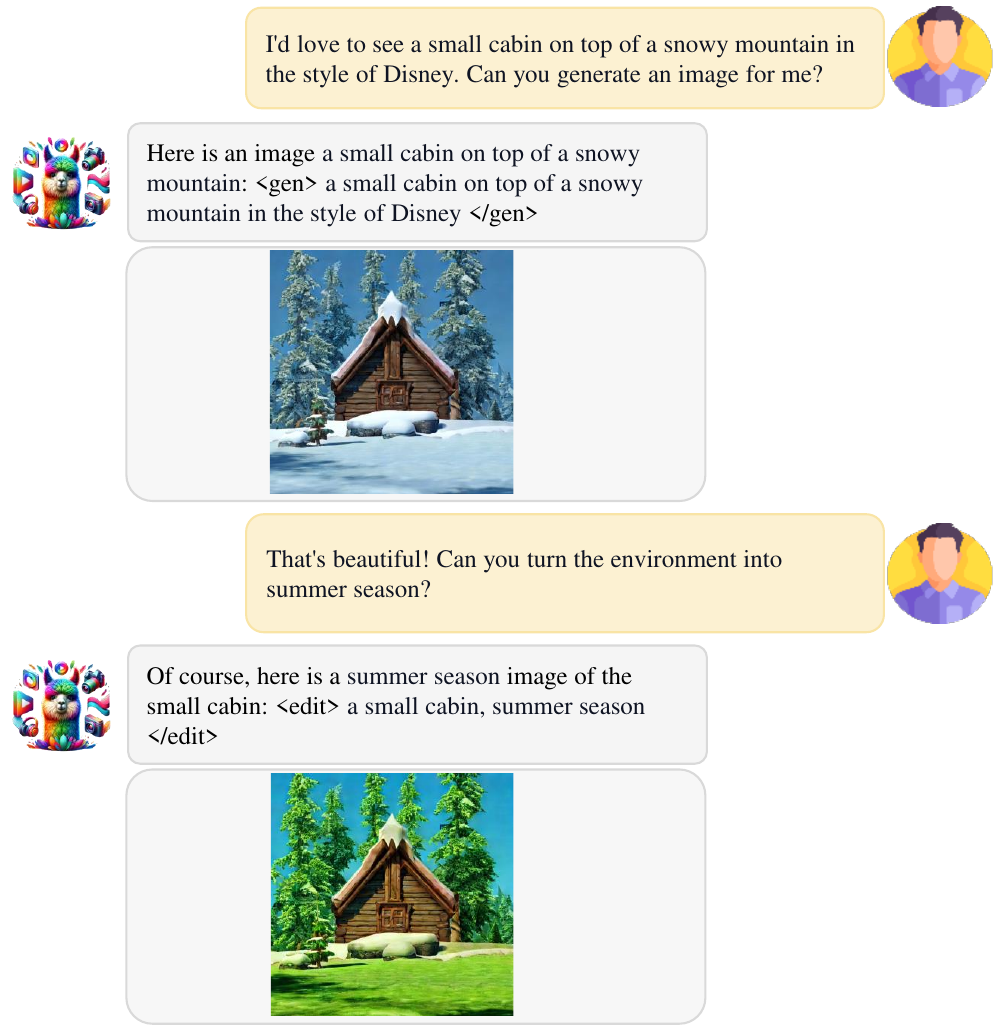}  
    \caption{\textbf{Example of Image Generation and Editing (4).}}  
    \label{fig:gen_edit_sample2}  
\end{figure}

  
\begin{figure}[htbp]  
    \centering  
    \includegraphics[width=1.0\textwidth]{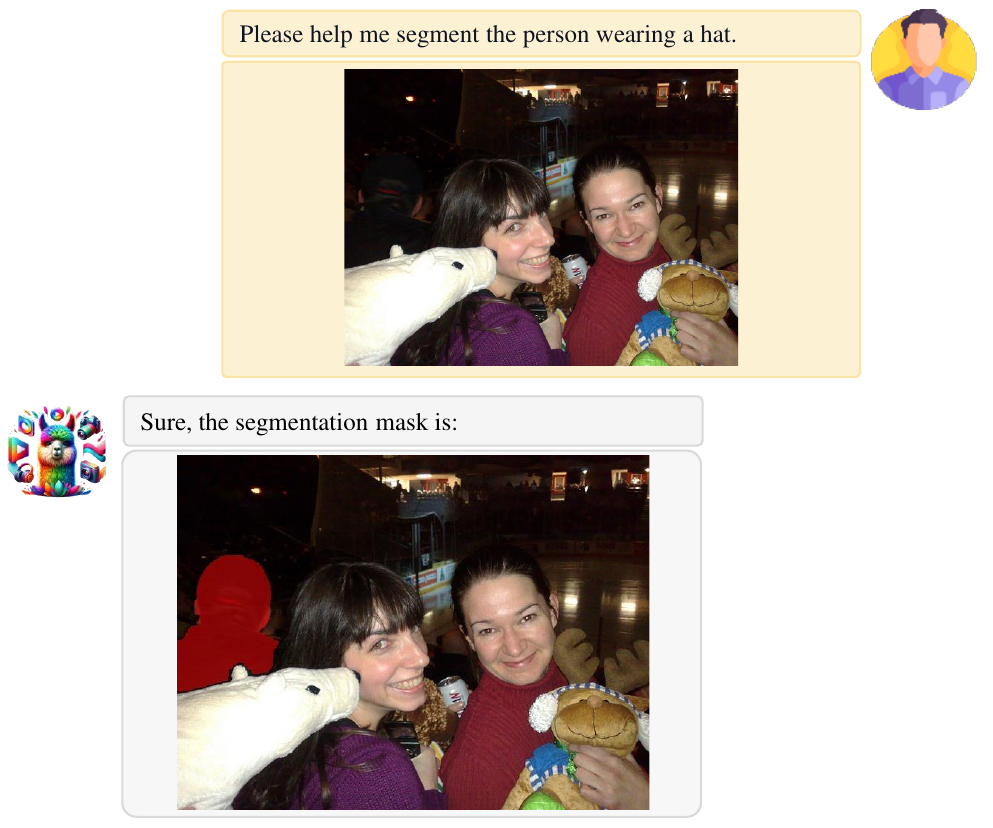}  
    \caption{\textbf{Example of Image Segmentation (1).}}  
    \label{fig:segmentation1}  
\end{figure}  
  
\begin{figure}[htbp]  
    \centering  
    \includegraphics[width=1.0\textwidth]{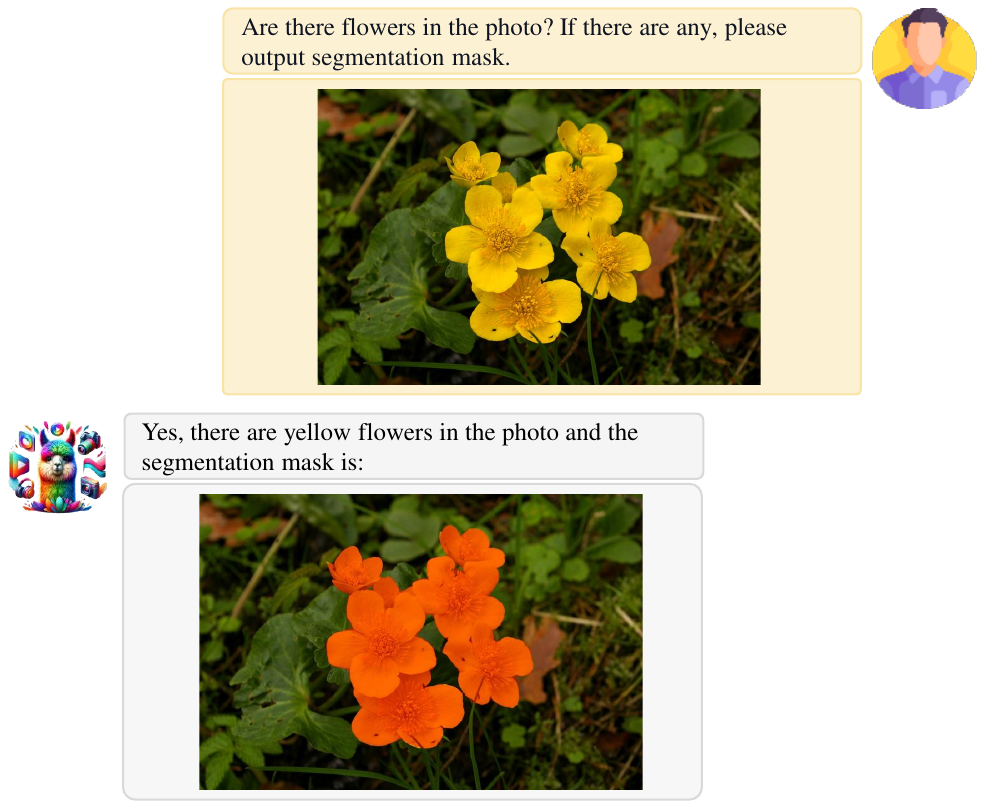}  
    \caption{\textbf{Example of Image Segmentation (2).}}  
    \label{fig:segmentation2}  
\end{figure}  
  
\begin{figure}[htbp]  
    \centering  
    \includegraphics[width=1.0\textwidth]{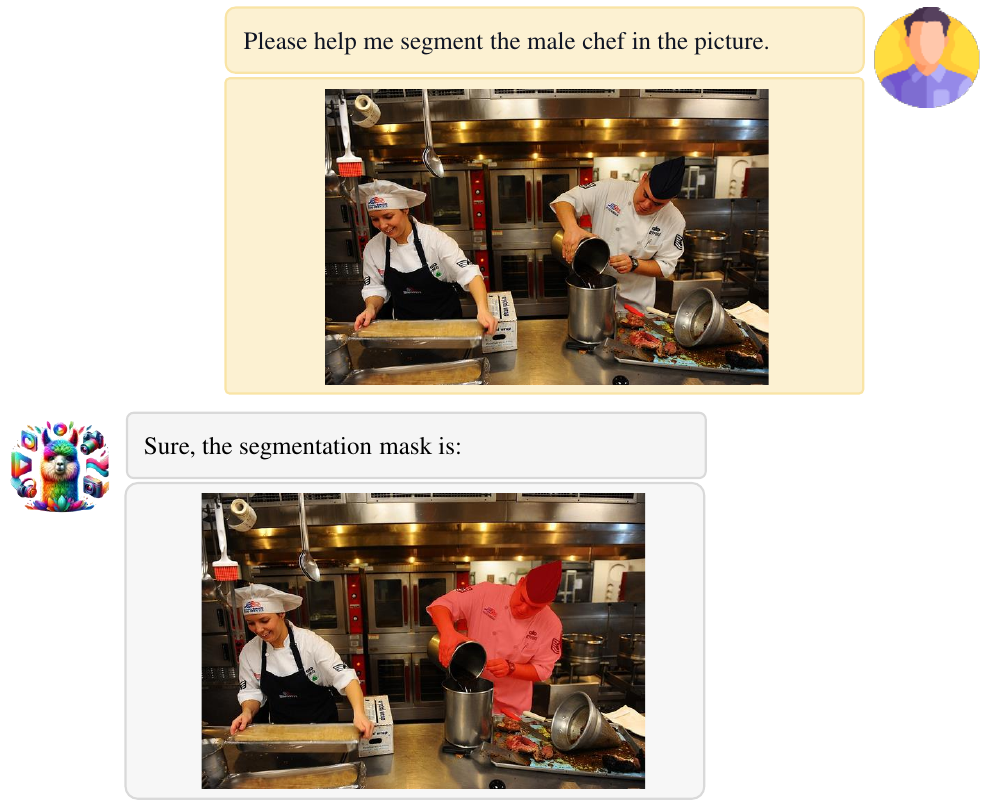}  
    \caption{\textbf{Example of Image Segmentation (3).}}  
    \label{fig:segmentation3}  
\end{figure}




\begin{figure}[htbp]
\centering
\includegraphics[width=1.0\columnwidth]{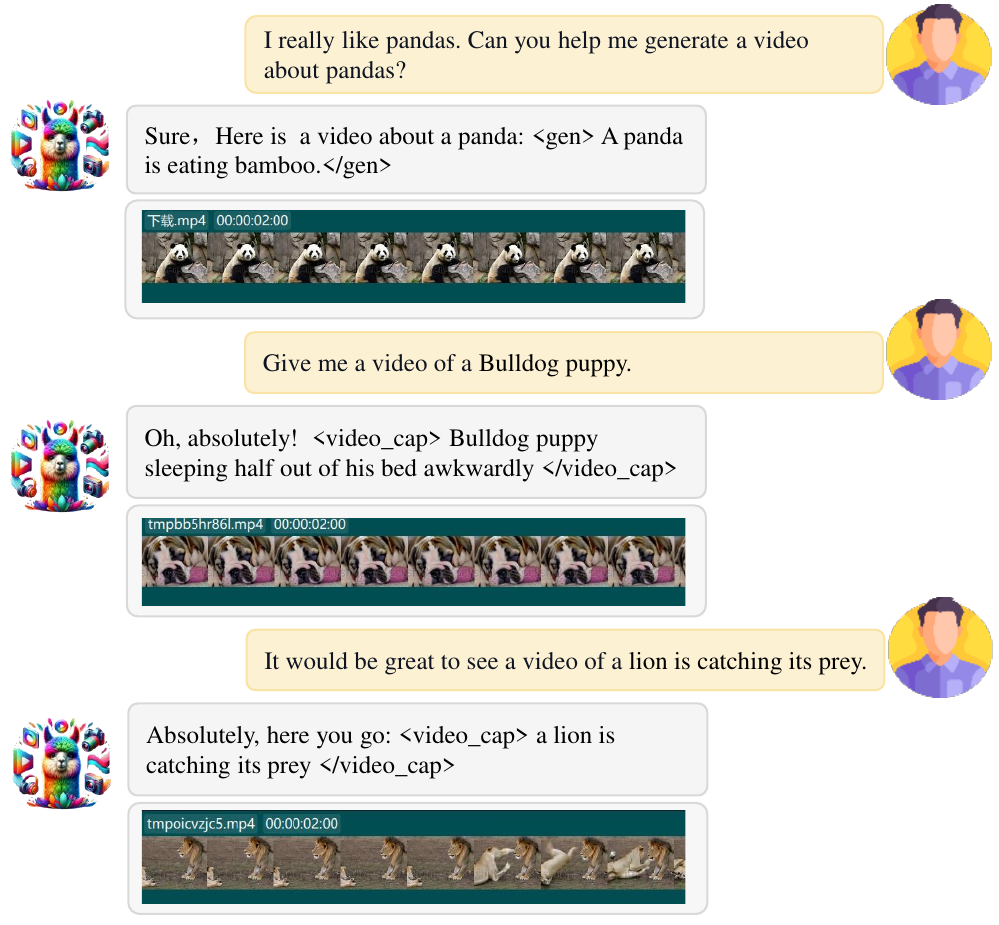}
\caption{\textbf{Examples of Video Generation.}}
\label{fig:appendix_vid_gen}
\end{figure}


\begin{figure}[htbp]
\centering
\includegraphics[width=1.0\columnwidth]{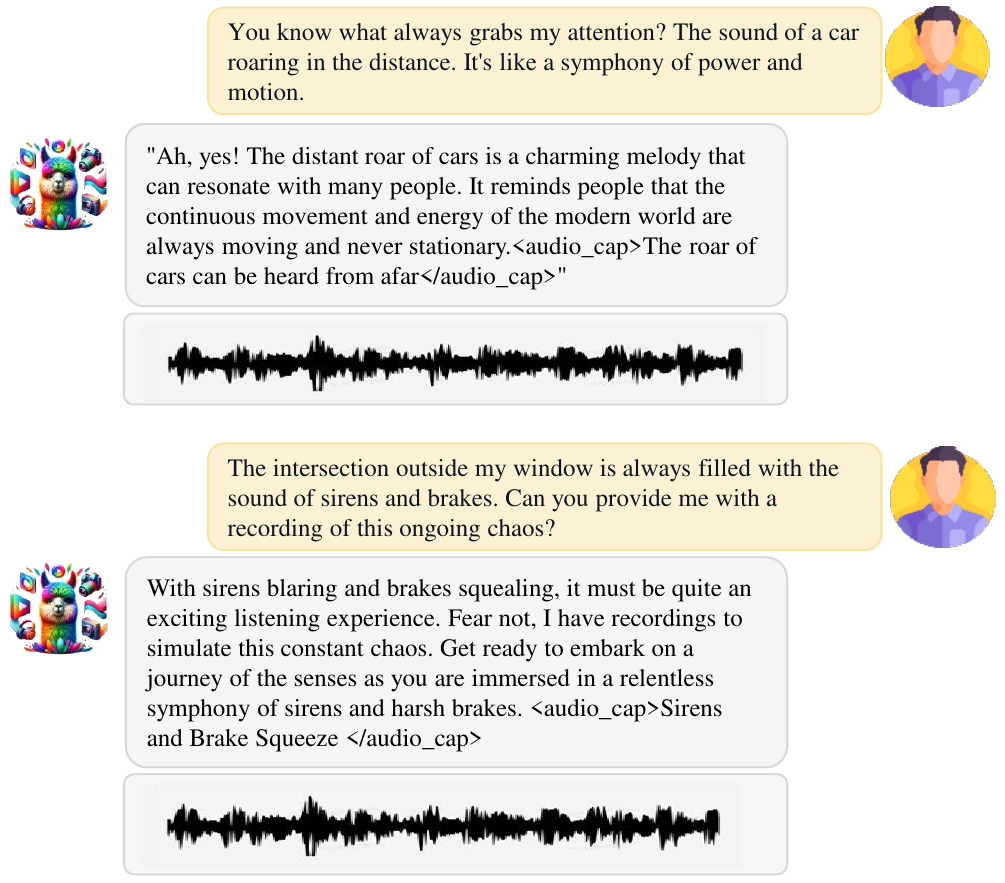}
\caption{\textbf{Examples of Audio Generation.}}
\label{fig:appendix_aud_gen}
\end{figure}



\begin{table*}[htbp]
\centering
\begin{tabular}{c|p{13cm}}
\midrule
\textbf{From} & \textbf{Value} \\
\hline
HUMAN  & I've been thinking about getting a tattoo of a cosmic jellyfish, can you help me visualize it? \\
\hline
GPT & Of course! Here's an image of a cosmic jellyfish: \redtoken{<gen>} a cosmic jellyfish with neon tentacles floating in space \redtoken{</gen>}
\\
\hline
HUMAN  & Wow, that's amazing! Can you show me a cosmic jellyfish with some galaxies in the background? \\
\hline
GPT & Certainly! Here's a modified image of the cosmic jellyfish with galaxies in the background: \redtoken{<edit>} a cosmic jellyfish with neon tentacles floating amidst colorful galaxies in space \redtoken{\redtoken{</edit>}} \\
\hline
HUMAN  & This is exactly what I had in mind! I'm so excited about this tattoo now. Can you show me a smaller version of the cosmic jellyfish, maybe in a minimalist style? \\
\hline
GPT & Absolutely! Here's a minimalist version of the cosmic jellyfish: \redtoken{<gen>} a simple line art of a cosmic jellyfish with a serene and modern aesthetic \redtoken{</gen>} \\
\hline
HUMAN  & I love it! It's going to be hard to choose which one to get. Thanks for all your help! \\
\hline
GPT & You're welcome! I'm glad I could assist you in visualizing your cosmic jellyfish tattoo. Let me know if there's anything else I can do for you. \\
\hline
\midrule

HUMAN  & I'm curious about a magical forest with glowing mushrooms. Can you show me what it looks like? \\
\hline
GPT & Certainly! Here is the concept of a magical forest with glowing mushrooms: \redtoken{<gen>} a magical forest with glowing mushrooms \redtoken{</gen>}
\\
\hline
HUMAN  & Wow, that looks amazing! Is there a special creature that lives in this forest? \\
\hline
GPT & Yes, there is! Let me show you a unique creature that inhabits this magical forest: \redtoken{<edit>} a magical forest with glowing mushrooms and a special creature \redtoken{</edit>} \\
\hline
HUMAN  & I'd love to see more details of the creature and its surroundings. Can you provide a closer look? \\
\hline
GPT & Absolutely! Here is a closer look at the special creature and its enchanting surroundings: \redtoken{<edit>} a magical forest with glowing mushrooms, featuring a special creature in its habitat \redtoken{</edit>} \\
\hline
HUMAN  & This is fantastic! Can you tell me more about the magical properties of the glowing mushrooms? \\
\hline
GPT & Certainly! The glowing mushrooms in this forest possess magical qualities, emitting a soft and enchanting light that illuminates the entire area, creating an otherworldly atmosphere. \\
\hline
\midrule

HUMAN  & I'm planning a trip to the beach; can you show me what a picturesque beach looks like? \\
\hline
GPT & Sure! Here is a breathtaking beach scene. \redtoken{<gen>} a picturesque beach with golden sand and crystal-clear blue water \redtoken{</gen>}
\\
\hline
HUMAN  & Wow, that looks amazing! Can you also show me a beach with palm trees and hammocks? \\
\hline
GPT & Absolutely! Here is a serene beach setting with palm trees and hammocks. \redtoken{<edit>} a picturesque beach with palm trees and hammocks \redtoken{</edit>} \\
\hline
HUMAN  & This looks so relaxing. I wonder what a beach at sunset looks like. Can you show me that? \\
\hline
GPT & Of course! Here is a stunning sunset beach view. \redtoken{<edit>} a picturesque beach at sunset \redtoken{</edit>} \\
\hline
HUMAN  & Oh, that's so beautiful. I'd love to have this as a desktop wallpaper. Can you create one for me? \\
\hline
GPT & Sure thing! Here is a desktop wallpaper featuring a picturesque beach at sunset. \redtoken{<edit>} a desktop wallpaper of a picturesque beach at sunset \redtoken{</edit>} \\
\hline
\midrule
\end{tabular}
\caption{\textbf{Interactive image generation and editing data}}
\label{tab:Interactive_Image_generation_and_editing_data}
\end{table*}

\begin{table*}[h]
\centering
\begin{tabular}{c|p{13cm}}
\midrule
\textbf{From} & \textbf{Value} \\
\hline
\midrule
HUMAN  & Can you find me an audio recording that showcases the progression of a motor's sound? I'm particularly interested in capturing the transition from a humming idle to a revving acceleration. \\
\hline
GPT & Certainly! I have just the audio clip you're looking for. It begins with the calm hum of the motor as it idles, and then gradually builds up to a roaring acceleration. This recording beautifully captures the evolving soundscape of the motor's performance. \redtoken{\textless audio\_cap\textgreater} A motor hums as it idles then accelerates \redtoken{\textless /audio\_cap\textgreater} \\
\hline
\midrule
HUMAN  & I'm in need of an audio file that captures the thrilling experience of an engine revving, with the added excitement of humming and squealing tires. Can you help me with that? \\
\hline
\hline
GPT & Absolutely! I've got the perfect audio snippet for you. It's a recording that captures the adrenaline-pumping sound of an engine revving, accompanied by the intense humming and the screeching of tires. Get ready to feel the rush! \redtoken{\textless audio\_cap\textgreater} Revving of an engine with humming and squealing tires \redtoken{\textless /audio\_cap\textgreater} \\
\hline
\midrule
HUMAN  & Can you create an audio clip with the sound of people speaking while a car engine runs? \\
\hline
GPT & Absolutely! I have just the audio clip you're looking for. It captures the ambient noise of a lively conversation happening alongside the constant hum of a running car engine. The combination of human voices and the mechanical sound of the engine creates an intriguing auditory experience. Enjoy! \redtoken{\textless audio\_cap\textgreater} People speaking while car engine runs \redtoken{\textless /audio\_cap\textgreater} \\
\hline
\midrule
HUMAN  & I'd love to listen to a recording of a duck quacking loudly with insect noises in the background. \\
\hline
GPT & Absolutely! I have just the audio clip for you. Prepare to be immersed in nature's orchestra as you hear the distinct sound of a duck quacking loudly, accompanied by the gentle symphony of buzzing insects in the background. Enjoy the harmonious blend of these natural sounds! \redtoken{\textless audio\_cap\textgreater} A duck quacking loudly with insect noises in the background \redtoken{\textless /audio\_cap\textgreater} \\
\hline
\midrule
HUMAN  & I need some audio that captures the essence of a thunderstorm, with the added element of a vehicle in motion. Can you help me find it? \\
\hline
GPT & Of course! Prepare to be transported into the heart of a thunderous storm. I've got the perfect audio recording for you: the symphony of rain cascading from the sky, intermingled with the rolls of thunder, all while a vehicle races through the rain-soaked streets. Get ready to feel the power and energy of nature in this captivating audio experience. \redtoken{\textless audio\_cap\textgreater} Rain falls and thunder rolls while a vehicle runs \redtoken{\textless /audio\_cap\textgreater} \\
\hline
\midrule
HUMAN  & I'm trying to recreate a specific audio ambiance in my story. It involves a man speaking, and in the background, you can hear birds chirping intermittently. Additionally, there's a small machine that makes short whirring sounds. Can you help me paint this audio picture? \\
\hline
GPT & Absolutely! Picture a tranquil scene outdoors, where the delicate melodies of chirping birds dance in the air. Amidst this natural symphony, a man's voice resonates, creating an engaging narrative. Accompanying his speech are brief intervals of a small machine emitting gentle whirring sounds, adding a touch of mechanical presence to the serene environment. It's a unique audio blend that brings together the beauty of nature and human activity. \redtoken{\textless audio\_cap\textgreater} A man speaks and a small machine whirs shortly several times while birds chirp in the background \redtoken{\textless /audio\_cap\textgreater} \\
\hline
\midrule
HUMAN  & I'm in the mood for some soothing sounds. Can you find me an audio recording of water flowing in a stream? \\
\hline
GPT & Of course! I have just the audio you're looking for. Here's a recording that captures the tranquil melody of water gently flowing in a serene stream. Enjoy the peaceful ambiance! \redtoken{\textless audio\_cap\textgreater} Water is flowing in a stream \redtoken{\textless /audio\_cap\textgreater} \\
\hline
\end{tabular}
\caption{\textbf{Interactive audio generation data}}
\label{tab:Interactive_audio_generation_data}
\end{table*}

\begin{table*}[h]
\centering
\begin{tabular}{c|p{13cm}}
\midrule
\textbf{From} & \textbf{Value} \\
\hline
\midrule
HUMAN  & I'm curious to see a video where someone switches a light on and off repeatedly. Can you find one for me? \\
\hline
GPT & Absolutely! I found a video that captures the act of toggling a light switch on and off repeatedly. It should satisfy your curiosity and provide an interesting visual representation of this action. \redtoken{\textless video\_cap\textgreater} Turning light on and off \redtoken{\textless/video\_cap\textgreater} \\
\hline
\midrule
HUMAN  & I'm curious to see a video where someone switches a light on and off repeatedly. Can you find one for me? \\
\hline
GPT & Absolutely! I found a video that captures the act of toggling a light switch on and off repeatedly. It should satisfy your curiosity and provide an interesting visual representation of this action. \redtoken{\textless video\_cap\textgreater} Turning light on and off \redtoken{\textless/video\_cap\textgreater} \\
\hline
\midrule
HUMAN  & I have a dusty floor that needs cleaning. I wonder what it would look like if a vacuum cleaner cleaned the floor during the daytime. \\
\hline
GPT & Ah, I can certainly help you visualize that. Imagine a bright and sunny day with rays of sunlight streaming through the windows. The vacuum cleaner glides across the floor, sucking up the dust particles, leaving behind beautifully clean and clear surfaces. The sound of the vacuum hums in the background, indicating the effectiveness of its cleaning. Here's a video clip capturing the scene for you to get a better sense of it. \redtoken{\textless video\_cap\textgreater} Vacuum cleaner cleaning the floor with dust on it during daytime \redtoken{\textless/video\_cap\textgreater} \\
\hline
\midrule
HUMAN  & Show me a video that encapsulates the allure of Venice during its carnival season. I want to witness the joyful interactions between tourists and the carnival masks in Piazza San Marco. \\
\hline
GPT & Your wish is my command! Here's a captivating video that perfectly captures the magic of Venice's carnival. It showcases the enchanting atmosphere of Piazza San Marco, where tourists mingle with the carnival masks, creating moments of pure joy and excitement. \redtoken{\textless video\_cap\textgreater} Venice - carnival masks are photographed with tourists in piazza san marco \redtoken{\textless/video\_cap\textgreater} \\
\hline
\midrule
HUMAN  & I could really use some uplifting music right now. Can you find a video of a happy man in headphones, listening to cheerful music and dancing in an armchair in a cozy room at home? \\
\hline
GPT & Absolutely! Here's a video that perfectly captures the essence of joy and relaxation. Watch as a young man in casual wear, sitting in an armchair in a cozy room at night, enjoys his weekend by listening to cheerful music on his smartphone. His infectious happiness will surely brighten up your day. \redtoken{\textless video\_cap\textgreater} Happy man in headphones listens to cheerful music on smartphone sitting and dancing in armchair in cozy room at home at night. young male person in casual wear is relaxing on weekend. 4k static shot \redtoken{\textless/video\_cap\textgreater} \\
\hline
\midrule
HUMAN  & I'd like to see a video of cute adorable children having fun together. It would be great if they were isolated on a white background. \\
\hline
GPT & Absolutely! I have just the video in mind. Get ready to witness the joyous sight of cute and adorable children engaging in fun activities, all set against a pristine white background. Enjoy the heartwarming moments of pure happiness and laughter! \redtoken{\textless video\_cap\textgreater} Cute adorable children having fun together isolated on white background \redtoken{\textless/video\_cap\textgreater} \\
\hline
\midrule
HUMAN  & I'm in need of some visual tranquility. Is there a video that showcases a waterfall cascading over moss-covered rocks, with the camera gradually panning towards a lush green landscape? \\
\hline
GPT & Absolutely! I have the perfect video to soothe your soul. Watch as water gently trickles on mossy rocks beneath a captivating waterfall, creating a serene atmosphere. The camera slowly pans to the right, unveiling a breathtaking view of a lush and vibrant green landscape that will transport you to a state of pure tranquility. \redtoken{\textless video\_cap\textgreater} Water trickles on mossy rocks underneath a waterfall, panning right towards a lush green landscape \redtoken{\textless/video\_cap\textgreater} \\
\hline
\end{tabular}
\caption{\textbf{Interactive video generation data}}
\label{tab:Interactive_video_generation_data}
\end{table*}

\end{document}